\documentclass[10pt,twocolumn,letterpaper]{article}

\usepackage{wacv}
\usepackage{times}
\usepackage{epsfig}
\usepackage{graphicx}
\usepackage{amsmath}
\usepackage{amssymb}

\usepackage{booktabs}

\usepackage[usenames,dvipsnames]{color}
\usepackage{paralist}
\usepackage{subcaption}
\usepackage{adjustbox}
\usepackage{multirow}
\usepackage{xspace}
\usepackage{verbatim}
\usepackage{enumerate}
\usepackage{algorithm}
\usepackage{algorithmic}

\usepackage{mathtools}
\usepackage{tabulary}
\usepackage[table]{xcolor}
\usepackage{booktabs}
\usepackage{tabularx}
\usepackage{tabu}
\usepackage{longtable,tabu}
\usepackage{booktabs}
\usepackage{rotating}

\usepackage[nospace]{cite}
\usepackage{caption}
\usepackage{url}
\usepackage[normalem]{ulem}
\usepackage{bm}
\usepackage[pagebackref=true,breaklinks=true,letterpaper=true,colorlinks,bookmarks=false]{hyperref}

\def\CellR#1{\cellcolor{red!25}#1}
\def\CellB#1{\cellcolor{blue!25}#1}

\newcommand{\fig}{{Fig.}\@\xspace}
\newcommand{\tab}{{Table}\@\xspace}

\graphicspath{{./figures/}}

\wacvfinalcopy 

\def\wacvPaperID{124} 

\ifwacvfinal\fi
\begin{document}

\title
  {
  Multi-Camera Action Dataset for Cross-Camera Action Recognition Benchmarking
  }

\renewcommand\footnotemark{}

\author
	{
	Wenhui~Li$^1$, Yongkang~Wong$^2$, An-An~Liu$^1$, Yang~Li$^1$, Yu-Ting~Su$^1$, Mohan~Kankanhalli$^{2,3}$\thanks
		{
		\copyright 2017 IEEE. 
		Personal use of this material is permitted. 
		Permission from IEEE must be obtained for all other uses, 
		in any current or future media, 
		including reprinting/republishing this material for advertising or promotional purposes, 
		creating new collective works, 
		for resale or redistribution to servers or lists, 
		or reuse of any copyrighted component of this work in other works.
		}\\
	~\\
  $^1$School of Electronic Information Engineering, Tianjin University, China%
  \\
  $^2$Interactive \& Digital Media Institute, National University of Singapore, Singapore%
  \\
  $^3$School of Computing, National University of Singapore, Singapore%
	}

\maketitle 
\ifwacvfinal\thispagestyle{empty}\fi

\begin{abstract}

Action recognition has received increasing attention from the computer vision and machine learning communities in the last decade.
To enable the study of this problem,
there exist a vast number of action datasets,
which are recorded under controlled laboratory settings, real-world surveillance environments, or crawled from the Internet.
Apart from the ``in-the-wild'' datasets,
the training and test split of conventional datasets often possess similar environments conditions,
which leads to close to perfect performance on constrained datasets.
In this paper,
we introduce a new dataset,
namely Multi-Camera Action Dataset (MCAD),
which is designed to evaluate the open view classification problem under the surveillance environment.
In total, 
MCAD contains 14,298 action samples from 18 action categories,
which are performed by 20 subjects and independently recorded with 5 cameras.
Inspired by the well received evaluation approach on the LFW dataset,
we designed a standard evaluation protocol and benchmarked MCAD under several scenarios.
The benchmark shows that while an average of 85\% accuracy is achieved under the closed-view scenario,
the performance suffers from a significant drop under the cross-view scenario.
In the worst case scenario,
the performance of 10-fold cross validation drops from 87.0\% to 47.4\%. 

\end{abstract}

\section{Introduction}
\label{sec:introduction}

Human action recognition has received increasing attention from the computer vision and machine learning community in the past few decades~\cite{Chris_AVC_1998,Laptev_ICCV_2003,Dollar_VSPETS_2005,Wang_IJCV_2013,Schuldt_ICPR_2004,Liu_CVPR_2011,Jain_CVPR_2013,Zhang_JEET_2015,Xu_CVPR_2015,Sapienza_IJCV_2014,Zha_BMVC_2015}.
Its importance is greatly driven by applications,
such as human-computer interaction, action video indexing and retrieval, advanced video surveillance and so on.  

In the early action recognition research,
most of the research works were focused on the single-view learning problem.
These works mainly focused on the extraction of robust feature representation
(\eg~spatial features~\cite{Chris_AVC_1998}, 
spatio-temporal features~\cite{Laptev_ICCV_2003,Dollar_VSPETS_2005}, 
covariance descriptors~\cite{Wang_TIP_2007,Han_AFGR_2008}, 
trajectories-based descriptor~\cite{Wang_IJCV_2013}, \etc)
and classification methodology~\cite{Schuldt_ICPR_2004}. 
More recently,
semantic feature representations
(\ie~local action attributes)
were explored for improved action classification performance~\cite{Liu_CVPR_2011,Jain_CVPR_2013,Wang_ICCV_2013a,Zhang_JEET_2015}.
As the performance are saturating on the constrained datasets,
several works have focused on cross-view learning problem~\cite{Jiang_TC_2015,Gao_NC_2016,Farhadi_LNCS_2008} and cross-domain learning problem~\cite{Cui_TC_2014,Bach_ICML_2004,Duan_PAMI_2012}.

Cross-view learning aims to map features obtained from multiple views into a common feature space to handle the variations in visual appearance.
In the case where a new action category is given,
it can utilize the feature mapping model to perform action recognition between two different camera views.
On the other hand,
existing datasets often contain limited samples for each action category
(see \tab~\ref{tab:db_review}).
To address this issue,
cross-domain learning aims to leverage the small-scale data from target domain together with a large-scale data from an auxiliary domain to augment the generalization ability for model learning~\cite{Wang_TSMCB_2012}.

In the existing literature,
many datasets are often collected under single camera view~\cite{Messing_ICCV_2009,Gorelick_PAMI_2007} or multiple views with overlapped observation~\cite{Weinland_ICCV_2007,Liu_TC_2016,Liu_TC_2015}.
Hence, 
it is hard to systematically evaluate the robustness of action recognition algorithms on similar yet different backgrounds and captured environments.
On the other hand, 
the samples from large scale action recognition dataset collected from the Internet,
such as UCF101~\cite{Soomro_TR_2012},
consists of complex action captured from dynamic background environments.
This type of datasets is ideal for deep learning based approaches~\cite{Xu_CVPR_2015,Zha_BMVC_2015}.

\newcolumntype{L}[1]{>{\raggedright\let\newline\\\arraybackslash\hspace{0pt}}m{#1}}
\newcolumntype{C}[1]{>{\centering\let\newline\\\arraybackslash\hspace{0pt}}m{#1}}
\newcolumntype{R}[1]{>{\raggedleft\let\newline\\\arraybackslash\hspace{0pt}}m{#1}}

\begin{table*}[t]
	\centering
	\caption
		{
		Overview of existing action recognition dataset.
		\maltese~indicates that the camera views are partially overlapped. 
		}
	\label{tab:db_review}
	\vspace{-2ex}
	\resizebox{\textwidth}{!}{
	\begin{tabular}{| C{4ex}| L{19ex} || C{7ex} | C{7ex} | C{7ex} | C{7ex} |	C{24ex} || C{3ex} | C{3ex} || C{3ex} | C{3ex} | }
		\toprule
		  \begin{sideways} Dataset Type        \end{sideways}
		&  Dataset Name
		& \begin{sideways} No. of Subjects  \end{sideways}
		& \begin{sideways} No. of Actions   \end{sideways}
		& \begin{sideways} No. of Views     \end{sideways}
		& \begin{sideways} No. of Samples   \end{sideways}
		& Image Resolution
		& \begin{sideways} Single-View         \end{sideways}
		& \begin{sideways} Multi-View          \end{sideways}
		& \begin{sideways} Indoor              \end{sideways}
		& \begin{sideways} Outdoor             \end{sideways} \\\midrule
		\multirow{8}{*}{\centerline{\begin{sideways} Constrained \end{sideways}}}
			&ADL~\cite{Messing_ICCV_2009}         	 & ~~5 & ~~10  & 1 & ~~~~~150   &	$1,280 \times 720~~~~$									& \checkmark	&   					&	\checkmark	&             \\
	    &IXMAS~\cite{Weinland_ICCV_2007}         & 10  & ~~11  & 5 & ~~1,650    & $320 \times 240$ 												&        			& \checkmark  &	\checkmark  &             \\
			&KTH~\cite{Schuldt_ICPR_2004}       	   & 25  & ~~~~6 & 4 & ~~~~~600   & $160 \times 120$ 												& \checkmark  &             & \checkmark 	& \checkmark	\\
	    &MuHAVi~\cite{Singh_AVSS_2010}      	   & 14  & ~~17  & 8 & ~~1,904 	  & $704 \times 576$   											&        			& \checkmark  & \checkmark  &     				\\
			&MV-TJU~\cite{Liu_TC_2015}	      		   & 20  & ~~22  & 2 & ~~~~~600 	& $640 \times 480$    										&  						&	\checkmark  &	\checkmark  &             \\
	    &M$^2$I~\cite{Liu_TC_2016}         			 & 20  & ~~22  & 2 & ~~1,784 	  & $320 \times 240$    										&  						&	\checkmark  &	\checkmark  &        			\\
	    &TUM Breakfast~\cite{Kuehne_CVPR_2014}   & ~~- & ~~10  & -	& ~~1,989 	& $320 \times 240$ 												& \checkmark  &             &	\checkmark  &      				\\
	    &Weizmann~\cite{Gorelick_PAMI_2007}      & ~~9 & ~~10  & 1 & ~~~~~~~90	& $180 \times 144$   											& \checkmark  &             &          		& \checkmark  \\ \midrule
		\multirow{10}{*}{\centerline{\begin{sideways} Consumer \end{sideways}}}
	    &ASLAN~~\cite{Kliper_PAMI_2012}          & ~~- & ~~~~8 & - & ~~~~~233 	& -~~ 																		&	\checkmark  &        			&	\checkmark  & \checkmark  \\
			&HMDB~\cite{Kuehne_ICCV_2014}            & ~~- & ~~51  & - &	~~6,766 	& -~~ 																		&	\checkmark  &       			&	\checkmark  & \checkmark  \\
			&Hollywood~\cite{Laptev_CVPR_2008}       & ~~- & ~~~~8 & - & ~~~~~233	  & -~~ 																		&	\checkmark  &         		&	\checkmark  & \checkmark  \\
			&Hollywood2~\cite{Marszalek_CVPR_2009}   & ~~- & ~~12  & - & ~~3,669	  & -~~ 																		&	\checkmark  &            	&	\checkmark  & \checkmark  \\
			&Olympic sports~\cite{Niebles_LNCS_2010} & ~~- & ~~16  & - & ~~~~~800 	&	-~~ 																		& \checkmark  &            	&	\checkmark  & \checkmark  \\
			&Standford 40~\cite{Yao_ICCV_2011}       & ~~- & ~~12  & - & ~~3,669	  & -~~ 																		& \checkmark  &          		&	\checkmark  & \checkmark  \\
			&UCF11~\cite{Liu_CVPR_2009}              & ~~- & ~~11  & - & ~~3,040 	  & -~~ 																		& \checkmark  &             &	\checkmark  & \checkmark  \\
			&UCF50~\cite{Rodriguez_CVPR_2008}        & ~~- & ~~50  & - & ~~6,676	  & -~~  																		& \checkmark  &             &	\checkmark  & \checkmark  \\
			&UCF101~\cite{Soomro_TR_2012}            & ~~- & 101 	 & - & 13,320 	  & -~~  																		& \checkmark  &             &	\checkmark  & \checkmark  \\
			&UCF Sports ~\cite{Rodriguez_CVPR_2008}  & ~~- & ~~10  & - &	~~~~~184 	& $720 \times 480$    										& \checkmark  &             &	\checkmark  & \checkmark  \\\midrule
		\multirow{7}{*}{\centerline{\begin{sideways} Surveillance \end{sideways}}}
			&MSR~\cite{yan_iccv_2007}            	   & 10  & ~~~~3 & 2 & ~~~~~~~~~- & $320 \times 240$ 												& \checkmark  &             &	\checkmark  &	\checkmark  \\
			&iLIDS~\cite{Over_TRECVID_2014}    	     & ~~- & ~~~~7 & 5 & ~~~~~~~~~-	& $720 \times 576$ 												&          		& \maltese  	& \checkmark	&             \\
			&UCF Aerial~\cite{ucf_aerial_2011}       & ~~- & ~~~~9 & - & ~~~~~~~~~-	& -~~  																		& \checkmark  &           	& 			     	& \checkmark  \\
			&UCF-ARG~\cite{ucf_arg_2011}             & 12  & ~~10  & 3 & ~~1,440 	  & $1,920 \times 1,080$										&   					& \checkmark  &            	& \checkmark  \\
			&UT-Interaction~\cite{ryoo_iccv_2009}    & ~~6 & ~~~~3 & - & ~~~~~160 	& $720 \times 480$ 												& \checkmark  &           	&            	&	\checkmark  \\
			& MCAD (Proposed)           						 & 20  & ~~18	 & 5 & 14,298 	  & $1,280 \times 960$ \& $704 \times 576$  &	  					& \maltese  	& \checkmark 	&             \\
		\bottomrule
	\end{tabular}
	}
\end{table*}

Based on the above discussions,
it is timely to have independently recorded multi-view constrained datasets, 
which provide standardized evaluation configuration to analyze the robustness of an action recognition system under unseen views.
In this paper, 
we present a new Multi-Camera Action Dataset (MCAD),
which consists of actions recorded with two types of CCTV cameras.
Each camera has similar but slightly different FOVs, view perspective, image resolution, and background.
The actions were independently performed on each camera view.
Benchmark performance with single-view state-of-the-art algorithms indicate that this dataset is very challenging,
especially for micro actions (\ie~action with small amount of motion area) and the cross-view action recognition scenario.
 
The rest of the paper is organized as follows.
Section~\ref{sec:db_review} reviews the existing datasets.
Section~\ref{sec:proposed_db} delineates the details of the proposed MCAD,
where the benchmark is discussed in Section~\ref{sec:benchmark}.
Section~\ref{sec:conclusion} concludes the paper.
\section{Dataset Review}
\label{sec:db_review}

\subsection{Constrained Datasets}

The constrained datasets are captured under controlled environments with constant background.
Most of them were recorded under the indoor environment,
which exhibited stable illumination conditions, 
fixed distance between person and cameras, 
and fixed direction of the actions.

\begin{figure}[!t]
	\centering
	\begin{minipage}{1.0\columnwidth}				
		\begin{minipage}{1.0\columnwidth}
			\begin{minipage}{0.19\columnwidth} \centerline{\includegraphics[width=1.0\linewidth]{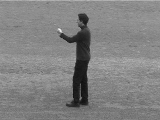}} \end{minipage} 
			\begin{minipage}{0.19\columnwidth} \centerline{\includegraphics[width=1.0\linewidth]{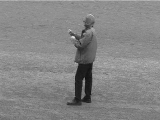}} \end{minipage}
			\begin{minipage}{0.19\columnwidth} \centerline{\includegraphics[width=1.0\linewidth]{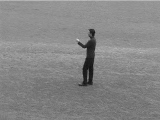}} \end{minipage} 
			\begin{minipage}{0.19\columnwidth} \centerline{\includegraphics[width=1.0\linewidth]{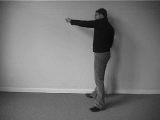}} \end{minipage} 
			\begin{minipage}{0.19\columnwidth} \centerline{\includegraphics[width=1.0\linewidth]{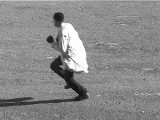}} \end{minipage}
		\end{minipage}
		\vspace{2pt}
				
		\begin{minipage}{1.0\columnwidth}
			\begin{minipage}{0.19\columnwidth} \centerline{\includegraphics[width=1.0\linewidth]{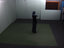}} \end{minipage} 
			\begin{minipage}{0.19\columnwidth} \centerline{\includegraphics[width=1.0\linewidth]{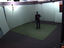}} \end{minipage}
			\begin{minipage}{0.19\columnwidth} \centerline{\includegraphics[width=1.0\linewidth]{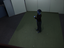}} \end{minipage} 
			\begin{minipage}{0.19\columnwidth} \centerline{\includegraphics[width=1.0\linewidth]{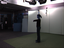}} \end{minipage} 
			\begin{minipage}{0.19\columnwidth} \centerline{\includegraphics[width=1.0\linewidth]{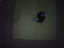}} \end{minipage}
		\end{minipage}
	\end{minipage}
	\vspace{-1ex}
	\caption
		{
		\small
		Sample images from constrained dataset.
		Top row: KTH dataset~\cite{Schuldt_ICPR_2004};
		Bottom row: IXMAS dataset~\cite{Weinland_ICCV_2007}.
		}
	\label{fig:review_constraints}
	\vspace{-2ex}
\end{figure}

The Weizmann dataset~\cite{Gorelick_PAMI_2007} contains clean and static background,
and the participants perform actions around a small area.
The KTH dataset~\cite{Schuldt_ICPR_2004} (see \fig~\ref{fig:review_constraints}) is considered more challenging than the Weizmann dataset.
It contains image sequences of human actions taken over from four scenarios and dynamic zoom variations.
The dataset consists of relatively simple actions,
such as ``walking'' and ``jump'', 
with limited action variations.
Literature has reported close to perfect performance on these datasets.
Specifically,
100\% classification accuracy on several action classes are reported in Weizmann dataset~\cite{Gorelick_PAMI_2007}.
Different from these actions,
there exist some datasets that recorded more complex actions.
In the Activity of Daily Living (ADL) dataset~\cite{Messing_ICCV_2009}, 
each activity is performed three times by five individuals of different shapes, size, gender, and ethnicity. 
Similarly,
the TUM Breakfast dataset~\cite{Kuehne_CVPR_2014} comprises of actions related to breakfast preparation in various kitchens.

As the performance on these databases is saturating, 
several cross-view action recognition datasets were proposed.
The first multi-view human action dataset is the INRIA Xmas Motion Acquisition Sequences (IXMAS) dataset~\cite{Weinland_ICCV_2007} (see \fig~\ref{fig:review_constraints}),
which contains actions taken from 5 calibrated and synchronized cameras (4 side views and 1 top view).
Subsequently,
the Multicamera Human Action Video (MuHAVi) dataset~\cite{Singh_AVSS_2010} collected multiple primitive actions video data using 8 CCTV cameras located at 4 sides and 4 corners of a rectangular platform.
Benefiting from the advances in depth sensing,
the MV-TJU dataset~\cite{Liu_TC_2015} contains actions performed in both light and dark environment from two different cameras.
Similarly,
the Multi-modal \& Multi-view \& Interactive (M$^2$I) dataset~\cite{Liu_TC_2016} extends the MV-TJU dataset by including person-person and person-object interactive action.
Both the MV-TJU and M$^2$I dataset consist of RGB image sequence, depth data and 3D skeleton data. 

As many reported results on the constrained datasets are very good,
these datasets are no longer regarded as challenging datasets for the action recognition problem. 
Furthermore, 
we argue that the actions are too simple when compared to the real world scenario.
The action samples in these datasets are synchronized in all cameras,
where the corresponding pairs have the same periodic properties. 
Several works are using this information to study the cross-view learning problem~\cite{Zheng_BMVC_2012} and cross-domain learning problem~\cite{Duan_PAMI_2012a,Hal_ACL_2007,Duan_PAMI_2012,Bach_ICML_2004}.
In addition, 
we note that the camera views employed in the training stage are unlikely to have direct relationship
(\ie~same view or overlapped region)
with the test camera,
especially for the surveillance application.

\subsection{Consumer generated Datasets} 

\begin{figure}[!t]
	\centering
	\begin{minipage}{1.0\columnwidth}
	   	\begin{minipage}{1.0\columnwidth}
	   		\begin{minipage}{0.242\columnwidth} \centerline{\includegraphics[width=1.0\linewidth]{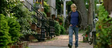}} \end{minipage} 
	   		\begin{minipage}{0.242\columnwidth} \centerline{\includegraphics[width=1.0\linewidth]{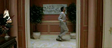}} \end{minipage} 
	   		\begin{minipage}{0.242\columnwidth} \centerline{\includegraphics[width=1.0\linewidth]{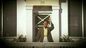}} \end{minipage} 
	   		\begin{minipage}{0.242\columnwidth} \centerline{\includegraphics[width=1.0\linewidth]{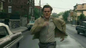}} \end{minipage} 
	   	\end{minipage}	   	
	   	\vspace{2pt}
	   	
     	 \begin{minipage}{1.0\columnwidth}
     	   	\begin{minipage}{0.242\columnwidth} \centerline{\includegraphics[width=1.0\linewidth]{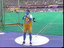}} \end{minipage} 
     	   	\begin{minipage}{0.242\columnwidth} \centerline{\includegraphics[width=1.0\linewidth]{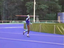}} \end{minipage} 
     	   	\begin{minipage}{0.242\columnwidth} \centerline{\includegraphics[width=1.0\linewidth]{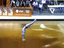}} \end{minipage} 
     	   	\begin{minipage}{0.242\columnwidth} \centerline{\includegraphics[width=1.0\linewidth]{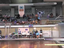}} \end{minipage}
     	 \end{minipage}
	\end{minipage}
	\vspace{-1ex}
	\caption
		{
		\small
		Sample images for consumer generated dataset.
		Top row: HMDB dataset~\cite{Kuehne_ICCV_2014};
		Bottom row: Olympic sports dataset~\cite{Niebles_LNCS_2010}.
		}
	\label{fig:review_consumer}
\end{figure}

The datasets of this category are generated by consumers and collected from the Internet, movies or personal video collections.
These datasets are very challenging when compared with constrained datasets,
due to its diversity in visual content, background complexity, and dynamic camera motion
Example of these datasets are shown in \fig~\ref{fig:review_consumer}.

University of Central Florida (UCF) has collected several challenging human action datasets.
UCF11~\cite{Liu_CVPR_2009},
UCF50~\cite{Rodriguez_CVPR_2008}, and
UCF101~\cite{Soomro_TR_2012} contain realistic videos and personal video collections collected from YouTube with different numbers of action classes.
UCF Sports Action~\cite{Rodriguez_CVPR_2008} consists of a set of actions in sports collected from a wide range of stock footage websites, 
including BBC Motion gallery and GettyImages. 
Other similar datasets include the Olympic sports dataset~\cite{Niebles_LNCS_2010}.
Moreover, 
the Human Motion Database (HMDB)~\cite{Kuehne_ICCV_2014} includes distinct action categories extracted from a wide range of sources. 
The Hollywood dataset~\cite{Laptev_CVPR_2008} and the Hollywood2 dataset~\cite{Marszalek_CVPR_2009} contain human actions distributed in the movies,
which enable the comprehensive benchmark for human action recognition in the realistic and challenging settings.
The Stanford 40 Action Dataset~\cite{Yao_ICCV_2011} contains images of humans performing 40 actions.
Different from these datasets which are designed for action classification problem,
Kliper-Gross~\etal~\cite{Kliper_PAMI_2012} proposed the Action Similarity Labeling (ASLAN) Challenge which contains 3697 action samples from 1571 unique YouTube videos divided into 432 non-trivial action categories.
This benchmark focuses on the action verification problem.  

\subsection{Surveillance Datasets}
 
\begin{figure}[!t]
	\centering
	\begin{minipage}{1.0\columnwidth}			
	   	\begin{minipage}{1.0\columnwidth}
	    	\begin{minipage}{0.242\columnwidth} \centerline{\includegraphics[width=1.0\linewidth]{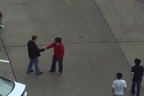}} \end{minipage} 
	    	\begin{minipage}{0.242\columnwidth} \centerline{\includegraphics[width=1.0\linewidth]{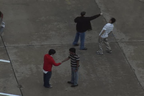}} \end{minipage} 
	    	\begin{minipage}{0.242\columnwidth} \centerline{\includegraphics[width=1.0\linewidth]{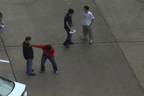}} \end{minipage} 
	    	\begin{minipage}{0.242\columnwidth} \centerline{\includegraphics[width=1.0\linewidth]{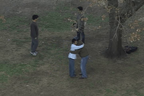}} \end{minipage} 
	    \end{minipage}
	   	\vspace{2pt}
	   	
     	 \begin{minipage}{1.0\columnwidth}
     	   	\begin{minipage}{0.242\columnwidth} \centerline{\includegraphics[width=1.0\linewidth]{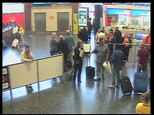}} \end{minipage} 
     	   	\begin{minipage}{0.242\columnwidth} \centerline{\includegraphics[width=1.0\linewidth]{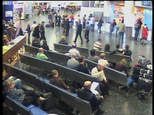}} \end{minipage} 
     	   	\begin{minipage}{0.242\columnwidth} \centerline{\includegraphics[width=1.0\linewidth]{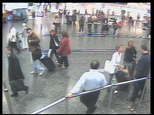}} \end{minipage} 
     	   	\begin{minipage}{0.242\columnwidth} \centerline{\includegraphics[width=1.0\linewidth]{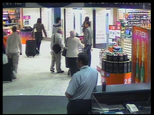}} \end{minipage}
     	 \end{minipage}
	\end{minipage}
	\vspace{-1ex}
	\caption
		{
		\small
		Sample images for surveillance dataset.
		Top row: UT-Interaction dataset~\cite{ryoo_iccv_2009}; 
		Bottom row: iLIDS dataset~\cite{Over_TRECVID_2014}.
		}
	\label{fig:review_surveillance}
\end{figure}
 
The dataset of this category is captured with fixed view cameras under the real-world surveillance environments,
which contains image sequences with complex background~\cite{Over_TRECVID_2014}, aerial view~\cite{ucf_aerial_2011}, and crowded unconstrained environment~\cite{Over_TRECVID_2014}.

The UCF Aerial Action dataset~\cite{ucf_aerial_2011} was obtained using a R/C-controlled blimp equipped with an HD camera mounted on a gimbal.
The collection represents a diverse pool of actions featured at different heights and various viewpoints.
The UT-Interaction dataset~\cite{ryoo_iccv_2009} focuses on human-human interactions in realistic environments in which each video contains at least one execution per interaction.
The MSR dataset~\cite{yan_iccv_2007} was created in 2009 to study the behavior recognition algorithms in presence of clutter and dynamic backgrounds and other types of action variations.
All the video sequences in this dataset are captured with clutter and moving backgrounds.
The UCF-ARG dataset~\cite{ucf_arg_2011} is a multi-view real-world dataset which consists of a ground camera, a rooftop camera, and an aerial camera mounted onto the payload platform of a helium balloon.
The iLIDS dataset~\cite{Over_TRECVID_2014} is another multi-view real-world dataset which collected action samples from indoor airport surveillance video in a busy airport.
This dataset is also used in the TRECVID Surveillance Event Detection (SED) evaluation since 2008,
where the presented action class remain challenging for the state-of-the-art approaches~\cite{Over_TRECVID_2014}.
\section{Multi-Camera Action Dataset}
\label{sec:proposed_db}

In this section, 
we delineate the details of the proposed dataset,
namely Multi-Camera Action Dataset (MCAD)\footnote{
available via http://mmas.comp.nus.edu.sg/MCAD/MCAD.html}.

\subsection{List of Recorded Actions}

The MCAD consists of 9 single person daily actions and 9 person-object actions.
These action categories are inherited from the KTH~\cite{Schuldt_ICPR_2004},
IXMAS~\cite{Weinland_ICCV_2007},
and iLIDS~\cite{Over_TRECVID_2014} datasets. 
The action list and respective definition of each action are shown in \tab~\ref{tab:action_list}.
Among these actions, 
there are 7 actions that contains action with small amount of motion area\footnote{
action ID: \{01, 02, 05, 10, 11, 12, \& 13\}},
we denoted these actions as micro action.
As demonstrated in Section~\ref{sec:benchmark},
these micro actions are more challenging,
especially the person-object actions.

In this dataset,
we recruited a total of 20 human subjects.
Each candidate repeats each action for 8 times
(4 times during the day and 4 times in the evening) 
under one camera view. 
Different from multi-view datasets such as IXMAS~\cite{Weinland_ICCV_2007} and MuHAVI~\cite{Singh_AVSS_2010}
where several cameras are deployed to record an action sample synchronously,
we use five cameras to record each action sample separately.
Therefore, 
an algorithm designed for cross-view learning problem that deliberately explores the properties across two simultaneously recorded action is not applicable.

\begin{figure*}[!h]
	\centering
	\begin{minipage}{2.0\columnwidth}
		\begin{minipage}{1.0\columnwidth}
			\begin{minipage}{0.02\columnwidth}	~		    						\end{minipage} 
	    \begin{minipage}{0.19\columnwidth}	\centerline{Cam04}	\end{minipage} 
	    \begin{minipage}{0.19\columnwidth}	\centerline{Cam05}	\end{minipage} 
			\begin{minipage}{0.19\columnwidth}	\centerline{Cam06}	\end{minipage} 
			\begin{minipage}{0.175\columnwidth}	\centerline{PTZ04}	\end{minipage} 
			\begin{minipage}{0.175\columnwidth}	\centerline{PTZ06}	\end{minipage}
	  \end{minipage}
	  
	  \vspace{2pt}
		\begin{minipage}{1.0\columnwidth}
	    \begin{minipage}{0.02\columnwidth}	\rotatebox{90}{Point}		    														\end{minipage} 
	    \begin{minipage}{0.19\columnwidth}	\centerline{\includegraphics[width=1.0\linewidth]{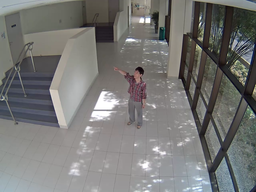}}	\end{minipage} 
	    \begin{minipage}{0.19\columnwidth}	\centerline{\includegraphics[width=1.0\linewidth]{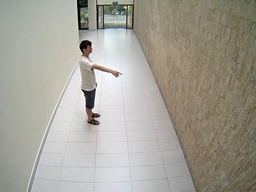}}	\end{minipage} 
			\begin{minipage}{0.19\columnwidth}	\centerline{\includegraphics[width=1.0\linewidth]{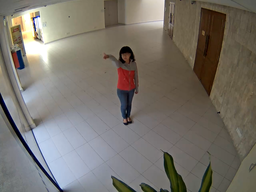}}	\end{minipage} 
			\begin{minipage}{0.175\columnwidth}	\centerline{\includegraphics[width=1.0\linewidth]{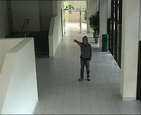}}	\end{minipage} 
			\begin{minipage}{0.175\columnwidth}	\centerline{\includegraphics[width=1.0\linewidth]{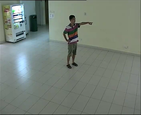}}	\end{minipage} 
		\end{minipage}
		
	  \vspace{2pt}
		\begin{minipage}{1.0\columnwidth}
	    \begin{minipage}{0.02\columnwidth}	\rotatebox{90}{Jump}		    														\end{minipage} 
	    \begin{minipage}{0.19\columnwidth}	\centerline{\includegraphics[width=1.0\linewidth]{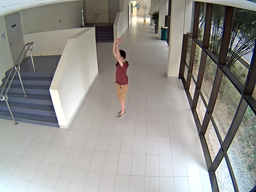}}	\end{minipage} 
	    \begin{minipage}{0.19\columnwidth}	\centerline{\includegraphics[width=1.0\linewidth]{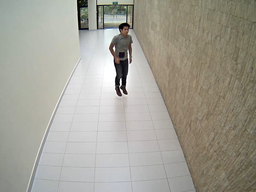}}	\end{minipage} 
			\begin{minipage}{0.19\columnwidth}	\centerline{\includegraphics[width=1.0\linewidth]{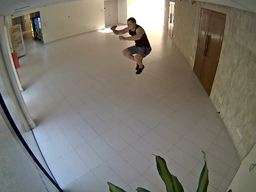}}	\end{minipage} 
			\begin{minipage}{0.175\columnwidth}	\centerline{\includegraphics[width=1.0\linewidth]{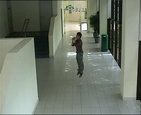}}	\end{minipage} 
			\begin{minipage}{0.175\columnwidth}	\centerline{\includegraphics[width=1.0\linewidth]{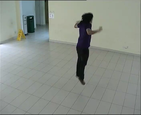}}	\end{minipage} 
		\end{minipage}
		
	  \vspace{2pt}
		\begin{minipage}{1.0\columnwidth}
	    \begin{minipage}{0.02\columnwidth}	\rotatebox{90}{CellToEar}		 														\end{minipage} 
	    \begin{minipage}{0.19\columnwidth}	\centerline{\includegraphics[width=1.0\linewidth]{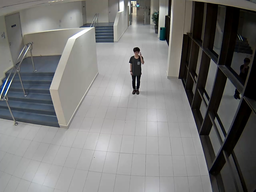}}	\end{minipage} 
	    \begin{minipage}{0.19\columnwidth}	\centerline{\includegraphics[width=1.0\linewidth]{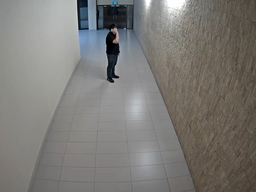}}	\end{minipage} 
			\begin{minipage}{0.19\columnwidth}	\centerline{\includegraphics[width=1.0\linewidth]{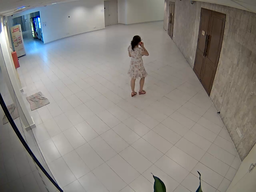}}	\end{minipage} 
			\begin{minipage}{0.175\columnwidth}	\centerline{\includegraphics[width=1.0\linewidth]{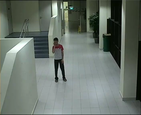}}	\end{minipage} 
			\begin{minipage}{0.175\columnwidth}	\centerline{\includegraphics[width=1.0\linewidth]{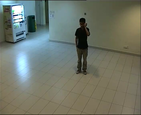}}	\end{minipage} 
		\end{minipage}
		
	  \vspace{2pt}
		\begin{minipage}{1.0\columnwidth}
	    \begin{minipage}{0.02\columnwidth}	\rotatebox{90}{ObjectThrow}  														\end{minipage} 
	    \begin{minipage}{0.19\columnwidth}	\centerline{\includegraphics[width=1.0\linewidth]{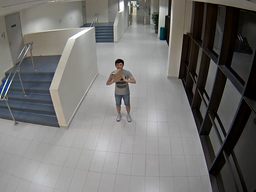}}	\end{minipage} 
	    \begin{minipage}{0.19\columnwidth}	\centerline{\includegraphics[width=1.0\linewidth]{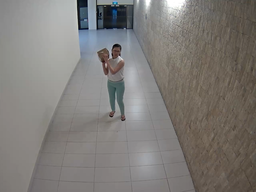}}	\end{minipage} 
			\begin{minipage}{0.19\columnwidth}	\centerline{\includegraphics[width=1.0\linewidth]{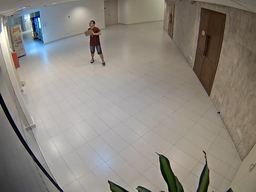}}	\end{minipage} 
			\begin{minipage}{0.175\columnwidth}	\centerline{\includegraphics[width=1.0\linewidth]{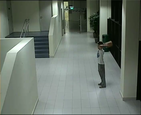}}	\end{minipage} 
			\begin{minipage}{0.175\columnwidth}	\centerline{\includegraphics[width=1.0\linewidth]{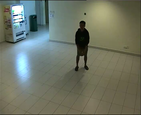}}	\end{minipage} 
		\end{minipage}
	\end{minipage}
	\caption
		{
		\small
		Sample images for the proposed MCAD dataset.
		Each row indicates unique action recorded with different indivduals on 5 distint camera views.
    Row 1 \& 2 show samples recorded during day time where images in row 3 \& 4 are recorded during night time.
		}
	\label{fig:action_sample}
\end{figure*}

During the recording stage,
we showed the subjects the list of actions and invited them to act freely with their personal preference.
As a result,  
not only we observed high intra action class variation among different action samples, 
we also noticed some individuals acted differently across different camera view or section (\ie~daytime or nighttime).
For example, 
the {\it Jump} action in \fig~\ref{fig:action_sample} demonstrates different posture on 5 randomly selected individuals.
Under all recordings, 
the individuals were allowed to face any direction within cameras' FOV.
This results in observable scale difference within the same camera view.
The only exception is PTZ06 where the corresponding FOV is narrower than other camera views.

\begin{table}[t]
	\centering
	\caption
		{
		\small
		List of actions and descriptions the proposed MCAD.
		Rows with {\textcolor{red!70}{RED}} and {\textcolor{blue!70}{BLUE}} background color indicate single-person action and person-object action,
		respectively.
		}
	\label{tab:action_list}
  \vspace{-2ex}
  \resizebox{\columnwidth}{!}{
	\begin{tabular}{c l l}
		\toprule
		ActionID		& Action Name   				& Action Description                              					\\	\midrule
		\CellR{01} 	& \CellR{Point}     		& \CellR{Someone points}                         						\\
		\CellR{02} 	& \CellR{Wave}      		& \CellR{Someone waves hand to catch peoples' attention}		\\
		\CellR{03} 	& \CellR{Jump}      		& \CellR{Someone jumps}                           					\\ 
		\CellR{04} 	& \CellR{Crouch}    		& \CellR{Someone crouches then stands up}         					\\
		\CellR{05}	& \CellR{Sneeze} 	  		& \CellR{Someone sneezes}                       						\\ 
		\CellR{06} 	& \CellR{SitDown}   		& \CellR{Someone sits down on a chair}        							\\
		\CellR{07} 	& \CellR{StandUp}   		& \CellR{Someone stands up from a chair}        						\\
		\CellR{08} 	& \CellR{Walk}   				& \CellR{Someone walks normally}                  					\\
		\CellR{09} 	& \CellR{PersonRun} 		& \CellR{Someone runs}                            					\\	\midrule
		\CellB{10} 	& \CellB{CellToEar}     & \CellB{Someone puts a cell phone to	his/her ear}  	      \\
		\CellB{11} 	& \CellB{UseCellphone}  & \CellB{Someone uses the cellphone to access information} 	\\ 		
		\CellB{12} 	& \CellB{DrinkingWater}	& \CellB{Someone uses a  bottle to drink water} 		 				\\
		\CellB{13} 	& \CellB{TakePicture}   & \CellB{Someone takes photos by using cellphone}           \\ 
		\CellB{14} 	& \CellB{ObjectGet}    	& \CellB{Someone bends or crouches to pick an object}       \\ 
		\CellB{15} 	& \CellB{ObjectPut}    	& \CellB{Someone puts down an object when walking}          \\ 
		\CellB{16} 	& \CellB{ObjectLeft}    & \CellB{Someone walks and drops an object in this process}	\\
		\CellB{17}	& \CellB{ObjectCarry}   & \CellB{Someone walks with a bag}                          \\ 	
		\CellB{18} 	& \CellB{ObjectThrow}   & \CellB{Someone throws a box to other place}               \\
		\bottomrule
	\end{tabular}
	}
\end{table}

\subsection{Environment Configuration}

The MCAD is recorded with five unique cameras, 
including three static cameras (\ie~Cam04, Cam05 \& Cam06) with fish eye
effect and two Pan-Tilt-Zoom~(PTZ) cameras (\ie~PTZ04 \& PTZ06).
These camera are mounted in a real-world surveillance environment.
Among these cameras, 
the Cam04-PTZ04 and Cam06-PTZ06 pairs covered the same region with different FOV.
The static camera has a resolution of {\small $1280 \times 960$} pixels.
The PTZ camera has a smaller FOV compared to the static camera,
where the image resolution is {\small $704 \times 576$} pixels.

The recording is carried out during both daytime and nighttime.
In all cases, 
though the actions are independently recorded for each camera, 
the illumination condition is constant for each individual.
However, 
due to the difference in the visual sensor and lens, 
we noticed observable difference in each camera view.
For example, 
the {\it ObjectThrow} samples showed in \fig~\ref{fig:action_sample} are all recorded during nighttime.
Although the lighting conditions are the same for all cameras, 
the recorded footage on PTZ06 appears to be darker than Cam06,
where both cameras observed the same region.

\subsection{Evaluation Metric}

In order to enable streamlined comparisons for future studies,
we adopt the evaluation protocol from the Labeled Faces in the Wild (LFW) dataset~\cite{LFW_Tech_2007}.
The MCAD is divided into two sets,
\ie~the Development Set and the Evaluation Set,
The Development Set is recommended for parameters tuning.
It consists of 10 randomly selected subjects from MCAD.
In this work, 
we use the Leave-One-Subject-Out Cross Validation (LOSOCV) strategy to evaluate the performance of an algorithm with various parameters.
The optimal parameters are then applied to the evaluation set for reporting results.
This protocol saves time during the comprehensive parameter search stage and creates an impartial condition for algorithm evaluation.

The Evaluation Set randomly divides all the subjects in MCAD into 10 training-test split\footnote{
NOTE: The data split is available from the MCAD website}. 
For each training-test split, 
12 subjects are selected as training data and the remaining 8 subjects as test set.
We report the final 10-fold cross validation result with estimated mean accuracy and the standard error of the mean as in~\cite{LFW_Tech_2007}.
Specifically,
the estimated mean accuracy {\small $\hat{\mu}$} is given by
\begin{equation}
	\hat{\mu} = \frac{\sum_{i=1}^{10} p_i}{10}
\label{eqn:result_mean}
\end{equation}

\noindent
where {\small $p_i$} is the accuracy from {\small $i$}-th fold.
The standard error of the mean is given as 
\begin{equation}
	S_{E} = \frac{\hat{\sigma}}{\sqrt{10}}
\label{eqn:result_std_err}
\end{equation}

\noindent
where {\small $\hat{\sigma}$} is the estimate of the standard deviation, 
given by
\begin{equation}
	\hat{\sigma} = \sqrt{ \frac{\sum_{i=1}^{10} \left( p_i - \hat{\mu} \right)^{2} }{9} }
\label{eqn:result_std_deb}
\end{equation}

\section{Benchmark}
\label{sec:benchmark}

\begin{figure*}[!t]
  \centering 
  \begin{minipage}{2.0\columnwidth}
  		\begin{minipage}{1.0\columnwidth}
	  		\begin{minipage}{0.19\columnwidth}	\centerline{\includegraphics[width=1.0\linewidth]{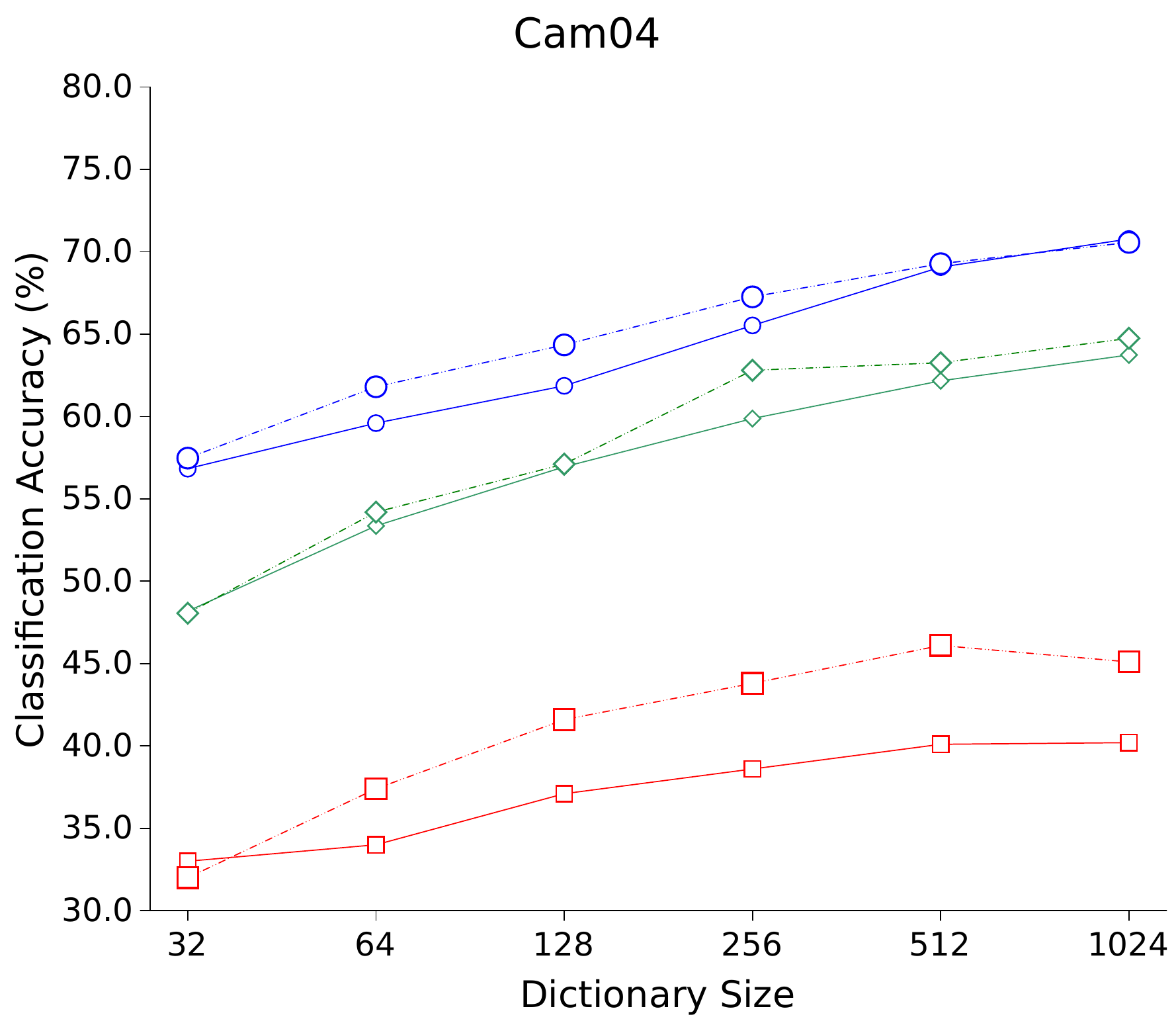}}			\end{minipage}
	  		\begin{minipage}{0.19\columnwidth}	\centerline{\includegraphics[width=1.0\linewidth]{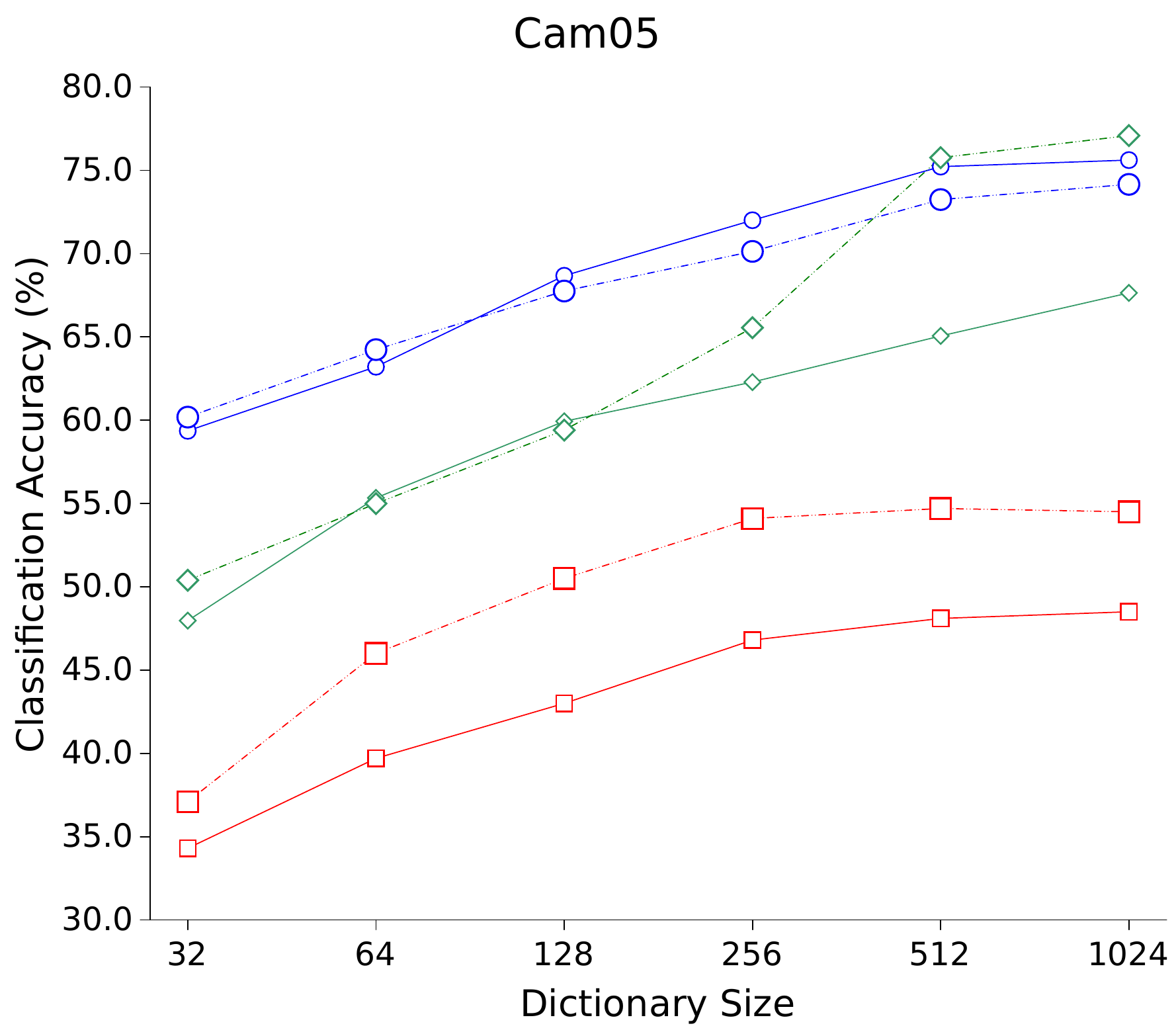}}			\end{minipage}
	  		\begin{minipage}{0.19\columnwidth}	\centerline{\includegraphics[width=1.0\linewidth]{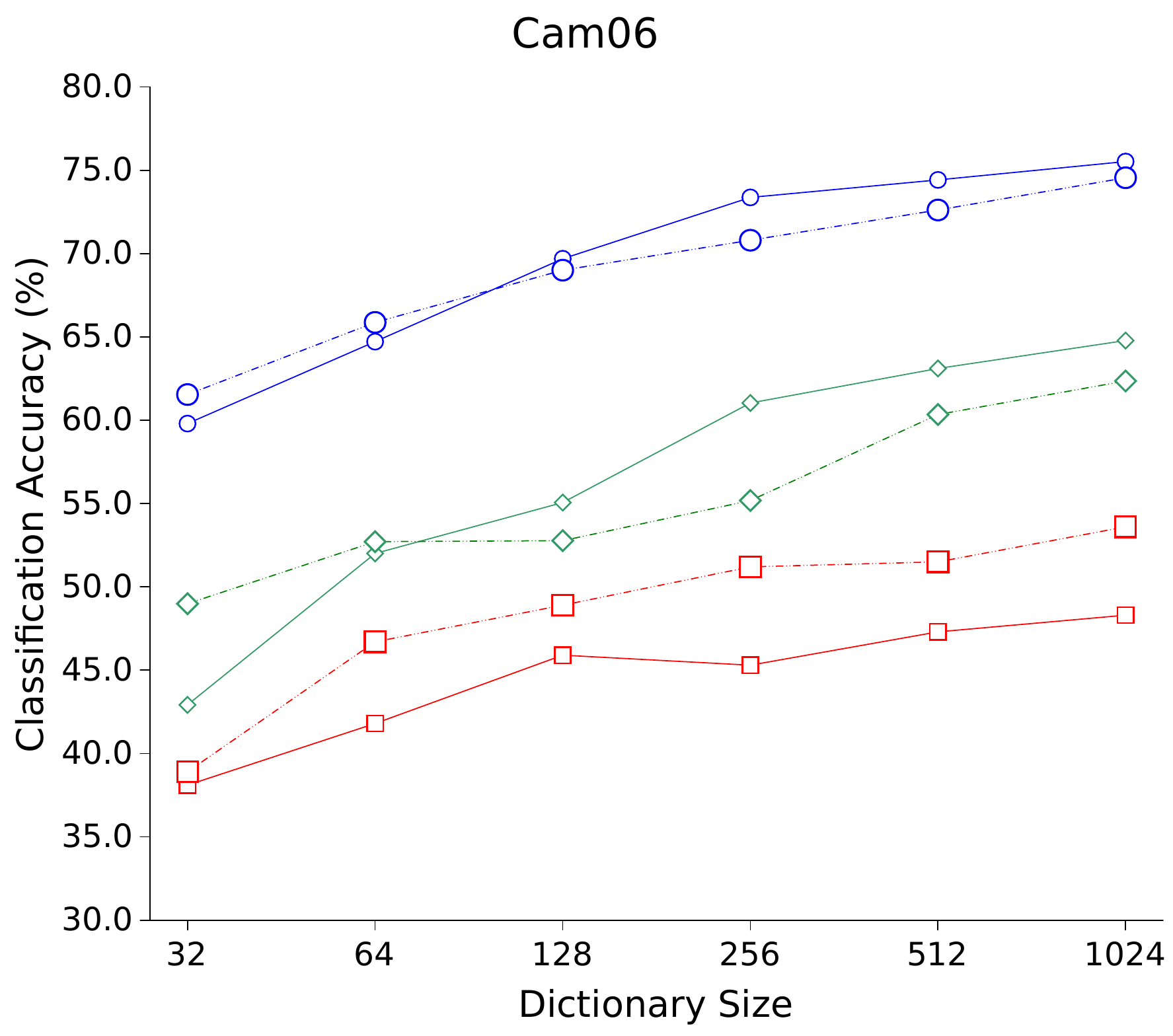}}			\end{minipage}
	  		\begin{minipage}{0.19\columnwidth}	\centerline{\includegraphics[width=1.0\linewidth]{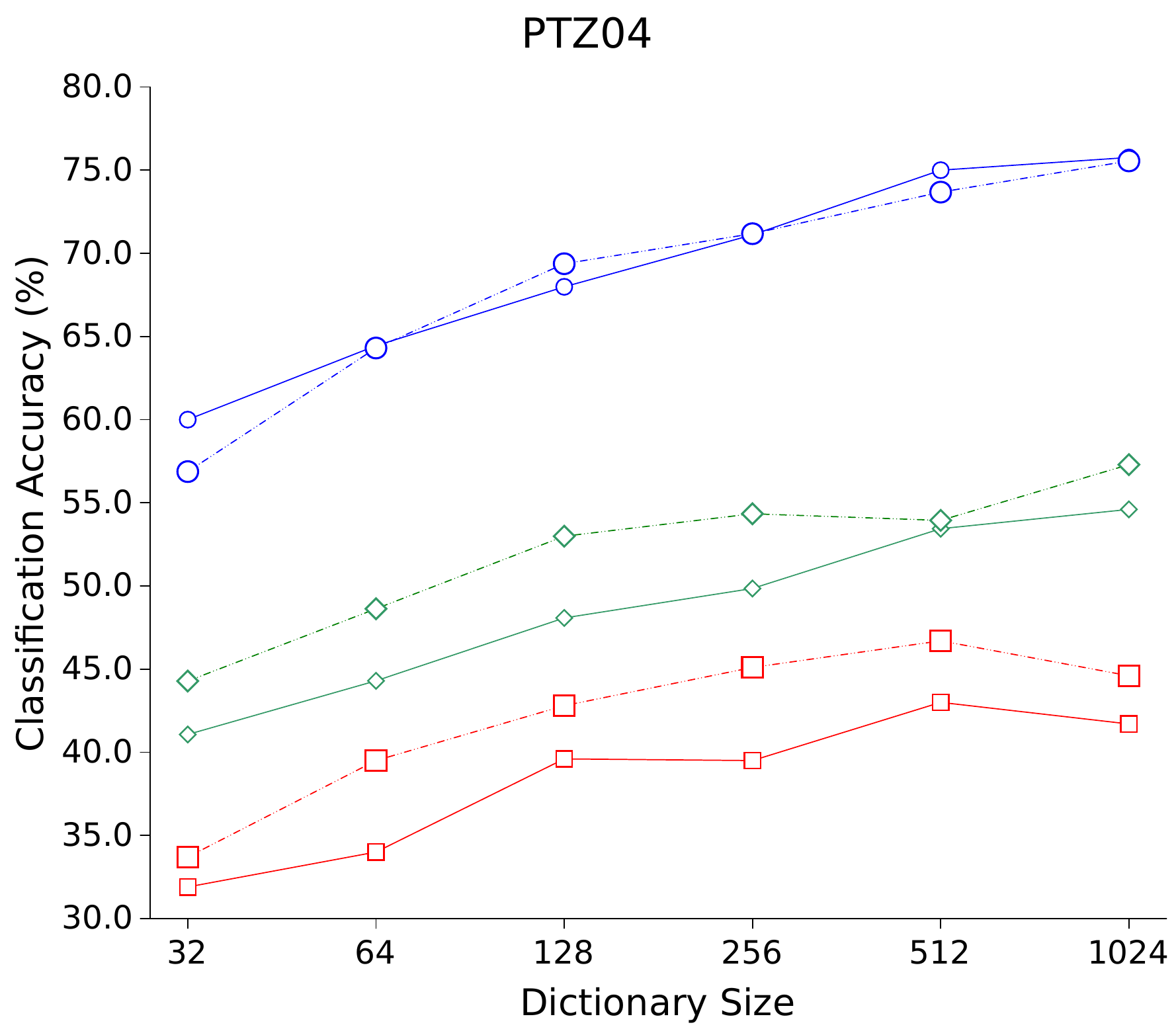}}			\end{minipage}
	  		\begin{minipage}{0.19\columnwidth}	\centerline{\includegraphics[width=1.0\linewidth]{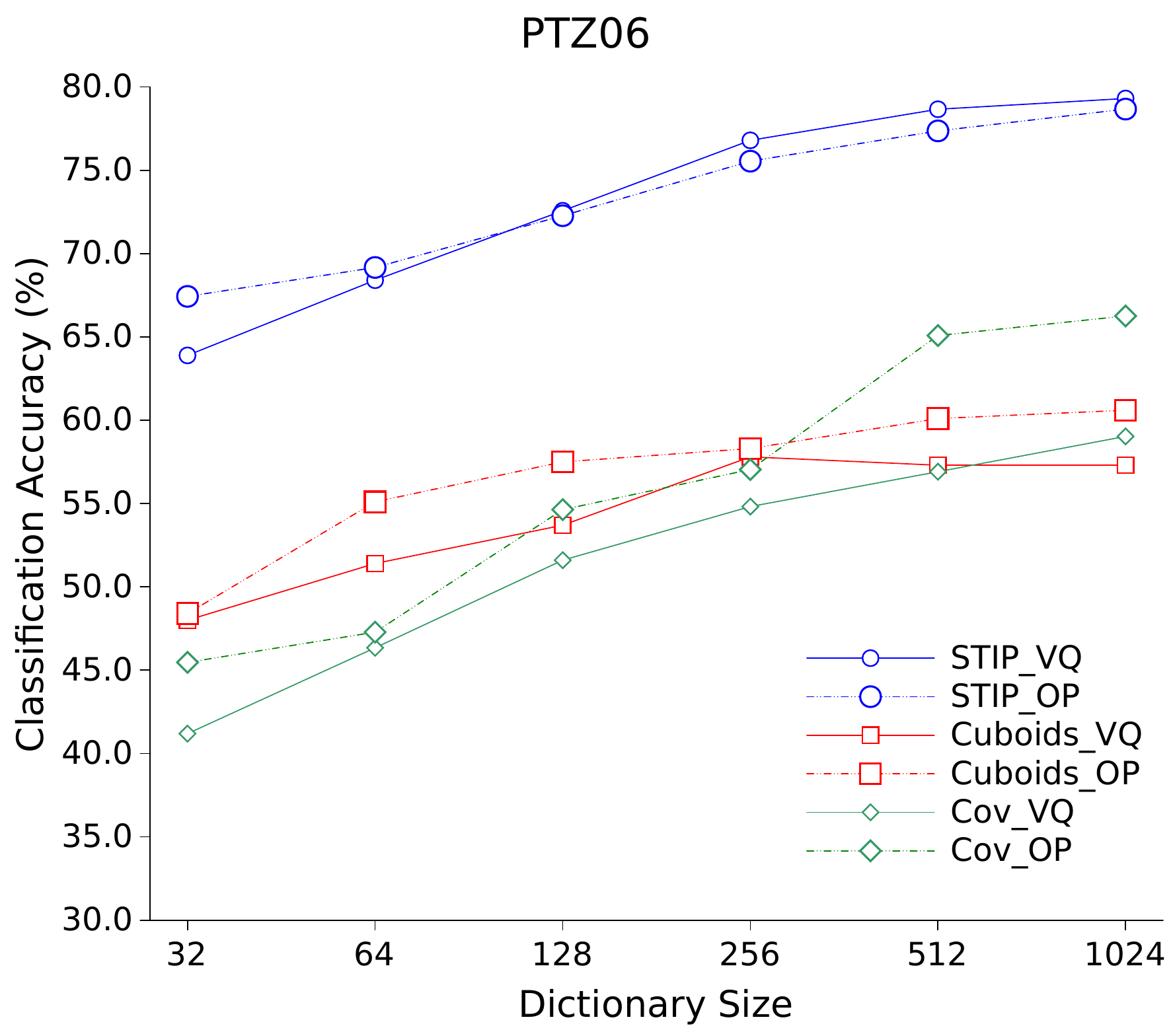}}			\end{minipage}
  		\end{minipage}
	\end{minipage}
	\vspace{-1ex}
	\caption
		{
		\small
		Performance of baseline algorithm on Development Set across various codebook size.
		}	
  \label{fig:dev_result}
\end{figure*}

\begin{table*}[!t]
	\centering
	\caption
		{
		\small
	  Closed-view classification accuracy (\%) on the Evaluation set.
	  Cells with \textcolor{blue}{BLUE} background color indicates the experiments are conducted with open-set classification constraints.
		}
	\label{tab:eval_result_SC}
	\vspace{-2ex}
	\resizebox{\textwidth}{!}{
	\begin{tabular}{|l|c|c|c|c|c|c|c|c|}
		\cmidrule{2-9}
		\multicolumn{1}{c}{}	& \multicolumn{2}{|c|}{STIP} 		& \multicolumn{2}{c|}{Cuboids}	&	\multicolumn{2}{c|}{Cov}			& \multicolumn{2}{c|}{IDT}						\\	\midrule
		Training \& Test Data	& VQ						& OP        		& VQ 						& OP        		& VQ            & OP     				& \multicolumn{2}{c|}{FV} 						\\	\midrule
		Cam04 								& $69.2\pm0.6$  &	$69.2\pm0.7$ 	& $40.9\pm0.6$  & $40.5\pm3.8$ 	& $61.3\pm0.4$	& $63.5\pm0.6$ 	& $88.6\pm0.5$ & \CellB{$83.5\pm0.6$} \\	
		Cam05 								& $74.0\pm0.6$  & $73.6\pm0.7$ 	& $47.7\pm0.4$  & $51.6\pm0.5$ 	& $65.4\pm0.6$ 	& $65.5\pm0.8$ 	& $91.6\pm0.3$ & \CellB{$86.2\pm0.4$} \\	
		Cam06 								& $72.5\pm0.4$  & $72.9\pm0.4$ 	& $48.0\pm0.7$  & $51.7\pm0.6$ 	& $63.3\pm0.4$ 	& $60.2\pm0.7$ 	& $90.1\pm0.3$ & \CellB{$83.6\pm0.9$} \\	
		PTZ04 								& $73.3\pm0.4$  & $73.4\pm0.5$ 	& $41.1\pm0.4$  & $45.5\pm0.5$ 	& $52.1\pm0.5$ 	& $55.2\pm0.5$ 	& $91.3\pm0.3$ & \CellB{$86.5\pm0.3$} \\	
		PTZ06 								& $77.1\pm0.6$  & $76.3\pm0.8$ 	& $54.9\pm0.8$  & $57.8\pm0.8$ 	& $56.9\pm0.5$ 	& $57.3\pm0.6$ 	& $91.3\pm0.7$ & \CellB{$87.0\pm0.6$} \\	\midrule
		All Static Cameras 		& $74.6\pm0.4$  &	$74.9\pm0.4$ 	& $48.7\pm0.3$  & $52.4\pm0.4$ 	& $65.1\pm0.4$ 	& $68.5\pm0.4$ 	& $92.8\pm0.2$ & \CellB{$84.2\pm0.9$} \\	
		All PTZ Cameras 			& $75.6\pm0.5$  &	$76.1\pm0.5$ 	& $48.0\pm0.6$  & $52.3\pm0.7$ 	& $54.7\pm0.5$	& $58.4\pm0.6$ 	& $92.3\pm0.2$ & \CellB{$87.2\pm0.5$} \\	\midrule
		All Cameras						& $75.6\pm0.4$	& $76.2\pm0.5$ 	&	$48.8\pm0.4$	& $53.7\pm0.6$ 	& $61.8\pm0.4$	& $66.1\pm0.3$ 	& $93.4\pm0.4$ & \CellB{$84.2\pm0.6$} \\
		\bottomrule
	\end{tabular}
	}
\end{table*}

\subsection{Baseline Algorithms}
\label{sec:benchmark_baseline}

In this work,
we benchmark the Bag-of-Words descriptor based method with four spatial-temporal local features and three encoding methods.
Specifically, 
we selected Spatio-Temporal Interest Point (STIP) feature~\cite{Laptev_ICCV_2003},
Cuboid feature~\cite{Dollar_VSPETS_2005},
Covariance matrices (denoted as Cov)~\cite{Faraki_CV_2014},
and Improved Dense Trajectory (IDT)~\cite{Wang_ICCV_2013}.
For Cuboid feature,
we use the parameter {\small $\sigma=2$} and {\small $\tau=1.5$} to extract up to 200 Cuboids from each action video,
followed by Principle Component Analysis (PCA)~\cite{Hastie_springer_2001} to reduce the dimensionality of the extracted feature to 100.
For the covariance matrices,
we first extract the 72-dimension HOF feature from Dense Trajectory (DT) feature to generate {\small $72 \times 72$} dimensional covariance matrices $X$.
Following~\cite{Faraki_CV_2014} we compute the Log-Euclidean vector representation of each $X$ and use this representation as covariance feature.

In our first set of baseline methods, 
we adopt two encoding methods for STIP, Cuboid and Cov features.
In the first encoding method, 
we utilized Kmeans++~\cite{Arthur_SDG_2007} to learn the codebook and use Vector Quantization to encode each local feature, 
followed by mean pooling to generate the descriptor.
We denote this baseline method as {\it Featurename}\_VQ.
For the second encoding method, 
we adopt a sparse coding approach,
where the K-SVD algorithm~\cite{Aharon_TSP_2006} is utilized for codebook learning and Orthogonal Matching Pursuit (OMP) algorithm~\cite{Tropp_TIT_2007} is used for encoding.
This baseline is denoted as {\it Featurename}\_OP.
For the IDT feature, 
we follow \cite{Wang_ICCV_2013} and use Fisher Vector (FV) encoding to generate the descriptor.
Specifically, 
we first use Gaussian Mixture Model (GMM) to learn the codebook.
Unlike VQ and OMP encoding method, 
FV encodes both the first and second order statistics between the video descriptors and a GMM.
We denote this baseline as IDT\_FV.

Given a descriptor,
we used an SVM with {\small $\chi^{2}$} kernel for classification.
The dimensionality of IDT\_FV descriptor is too high for {\small $\chi^{2}$} kernel SVM. 
We select a linear SVM,
which shows good results in classification for descriptor with high dimensionality~\cite{Fan_JMLR_2008,Douglas_springer_2009},
The aforementioned classification is deployed for closed-set classification scenario.
In this benchmark, 
we also employ an open-set linear SVM classifier~\cite{Bendale_CVPR_2015} to evaluate the performance under open-set scenario.
Based on the preliminary experiment, 
near and far plane pressures is fixed to 0.4 and 1.0,
respectively.

\begin{figure*}[!t]
	\centering 
		\centerline{\includegraphics[width=1.0\textwidth]{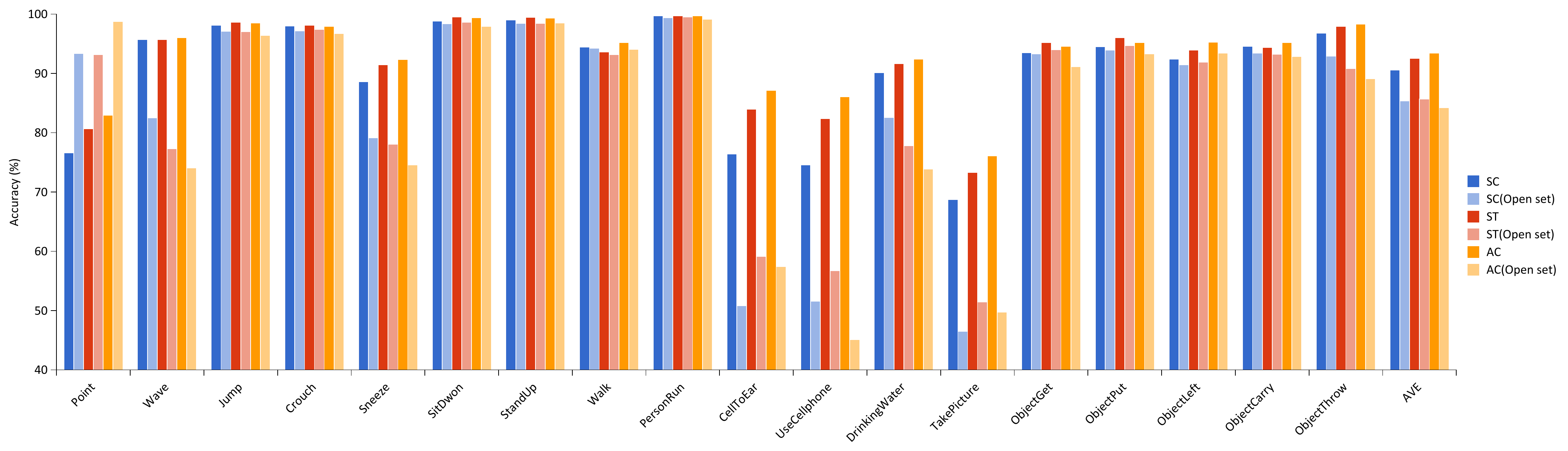}}
	\vspace{-2ex}
	\caption
		{   
		Mean classification accuracy of IDT\_FV descriptor on the Evaluation set under Single Camera (SC), Same Type (ST), and All Camera (AC) scenarios.
		}
	\label{fig:IDT_SC_3P}
\end{figure*}

\subsection{Evaluation under Closed View Scenario}
\label{sec:benchmark_closed_view}

In this section,
we evaluate the benchmark performance with closed view recognition scenario,
\ie~the camera view of the test data is the same as that for the training data.
Specifically, 
we restrict the training and test data from same camera source,
while the subjects can only appear in either the training or test data.
Three types of camera source scenarios are evaluated, 
namely Single Camera (SC),
Same camera Type (ST), and
All Cameras (AC).  

First, 
we use the Development Set to fine-tune the optimal codebook size and parameters of SVM classifier under closed-set classification scenario.
\fig~\ref{fig:dev_result} shows the performance on all 5 camera views.
Across all camera views,
the accuracy gradually increases and saturates when the codebook size is set to 1024.
The only exception is for the Cuboids feature based descriptor, 
which the performance under codebook size of 512 is the best.
Due to the limits of computational resources, 
the codebook size of IDT\_FV is evaluated up to 256.
Based on the optimal performance, 
we conducted 10-fold cross validation on the Evaluation Set.
The results are shown in \tab~\ref{tab:eval_result_SC} and the category wise performance of the best performing descriptor are shown in \fig~\ref{fig:IDT_SC_3P}. 
The key findings are as follows:
\begin{enumerate}
  \item 
  IDT\_FV consistently achieved the highest mean accuracy and lowest standard error on all scenarios,
  which is consistent with the reported performance on other datasets~\cite{Wang_ICCV_2013}.
  For the STIP and Cuboids features, 
  we notice that the performance with PTZ06 is better than for other camera views.
  As shown in \fig~\ref{fig:action_sample},
  the FOV of PTZ06 is narrower and the size of each person is more consistent than in other cameras.
  Hence, 
  the extracted local features is more consistent.
  \item 
  We observed that when the available training data increases from single camera view to all cameras,
  the performance of Cov\_OP and IDT\_FV increases.
  However, 
  this is not true for the VQ based encoding method, 
  where some of the single camera view scenarios report the best performance.
  This suggests that IDT\_FV not only is more discriminative than other baseline methods, 
  it also is more robust in handling training data with large environmental variation.
  \item 
  From \fig~\ref{fig:IDT_SC_3P}, 
  we found that the performance of micro actions, 
  such as {\it Sneeze}, {\it CellToEar}, and {\it TakePicture},
  is significant lower than the other class of actions.
  One reasons is that the available local features on the micro actions are fewer.
  On the other hand,
  action with large spatial movement, 
  such as {\it PersonRun} and {\it Jump},
  can be accurately recognized. 
  In generally, 
  person-object actions are slightly harder to recognize than single person actions.
\end{enumerate}

\begin{table*}[!t]
	\centering
	\caption
		{
		\small
    Open-view classification accuracy (\%) with IDT\_FV on the Evaluation set.
    First column indicates the source of the training data while the remaining columns are the evaluation with respective test image sequences. 
    The diagonal entries 
    (\ie~cells with \textcolor{red!80}{RED} background color) 
    are the classification accuracy of closed view SC scenario for comparison purposes.
    }
	\label{tab:eval_result_SC_CC_2_IDT}
	\vspace{-2ex}
	\resizebox{\textwidth}{!}{
	\begin{tabular}{|c|c|c|c|c|c||c|c|c|c|c|}
		\cmidrule{2-11}
		\multicolumn{1}{c|}{} & \multicolumn{5}{c||}{Closed-Set Classification}	& \multicolumn{5}{c|}{Open-Set Classification}							\\	\midrule
		Training Data	& Cam04 								&Cam05 									& Cam06 								& PTZ04 								& PTZ06
									& Cam04 								&Cam05 									& Cam06 								& PTZ04 								& PTZ06									\\	\midrule
		Cam04 				& \CellR{$88.6\pm0.6$} 	& $81.5\pm0.6$ 					& $75.6\pm0.6$ 					& $74.6\pm0.7$ 					& $63.4\pm0.7$  
									& \CellR{$83.5\pm0.6$} 	& $73.1\pm1.0$ 					& $64.1\pm1.1$ 					& $65.8\pm0.9$ 					& $52.9\pm1.3$ 					\\
		Cam05 				& $80.9\pm0.7$ 					& \CellR{$91.6\pm0.3$}  & $71.6\pm0.7$ 					& $73.9\pm0.4$ 					& $62.3\pm0.8$  
									& $75.2\pm0.8$ 					& \CellR{$86.2\pm0.4$}	& $59.8\pm0.8$ 					& $64.3\pm0.5$ 					& $51.6\pm1.2$ 					\\
	  Cam06 				& $73.0\pm0.6$ 					& $72.8\pm0.4$				 	& \CellR{$90.1\pm0.3$} 	& $65.8\pm0.4$ 					& $68.5\pm0.5$
									& $68.0\pm0.7$ 					& $65.1\pm0.9$					& \CellR{$83.6\pm0.9$} 	& $58.7\pm0.7$ 					& $58.1\pm1.2$ 					\\
    PTZ04 				& $75.7\pm0.4$ 					& $72.9\pm0.5$					& $67.8\pm0.9$ 					& \CellR{$91.3\pm0.3$} 	& $59.3\pm1.1$ 
									& $68.5\pm0.7$ 					& $63.7\pm0.7$					& $54.0\pm0.7$ 					& \CellR{$86.5\pm0.3$} 	& $47.4\pm1.0$ 					\\
    PTZ06 				& $57.4\pm0.6$ 					& $59.1\pm0.5$					& $60.3\pm0.4$ 					& $56.9\pm0.4$ 					& \CellR{$91.3\pm0.7$} 
									& $53.0\pm0.6$ 					& $53.5\pm0.5$					& $54.0\pm0.6$ 					& $50.2\pm0.6$ 					& \CellR{$87.0\pm0.6$} 	\\
    \bottomrule
	\end{tabular}
	}
	\vspace{-1.5ex}
\end{table*}

\begin{table*}[!t]
	\centering
	\caption
		{
		\small
    Open-view classification accuracy (\%) with IDT\_FV on the Evaluation set in IXMAS dataset.
    First column indicates the source of the training data while the remaining columns are the evaluation with respective test image sequences. 
    The diagonal entries 
    (\ie~cells with \textcolor{red!80}{RED} background color) 
    are the classification accuracy of closed view SC scenario for comparison purposes.
    }
	\label{tab:eval_result_SC_CC_2_IDT_IXMAS}
	\vspace{-2ex}
	\resizebox{\textwidth}{!}{
	\begin{tabular}{|c|c|c|c|c|c||c|c|c|c|c|}
		\cmidrule{2-11}
		\multicolumn{1}{c|}{} & \multicolumn{5}{c||}{Closed-Set Classification}	& \multicolumn{5}{c|}{Open-Set Classification}						\\	\midrule
		Training Data	& Cam0 								&Cam1 									& Cam2 								& Cam3 										& Cam4
									& Cam0 								&Cam1 									& Cam2 								& Cam3 										& Cam4									\\	\midrule
		Cam0 			& \CellR{$95.6\pm0.9$} 	& $87.0\pm2.2$ 					& $48.1\pm2.2$ 					& $64.7\pm2.6$ 					& -  
							& \CellR{$88.6\pm2.1$} 	& $74.0\pm2.2$ 					& $47.0\pm1.9$ 					& $60.0\pm2.9$ 					& - 										\\
		Cam1 			& $71.2\pm2.2$ 					& \CellR{$95.8\pm1.0$}  & $39.0\pm2.2$ 					& $43.6\pm2.2$ 					& -  
							& $60.5\pm4.2$ 					& \CellR{$94.8\pm1.2$}	& $36.1\pm3.3$ 					& $36.6\pm4.7$ 					& -											\\
	  Cam2 			& $71.2\pm2.5$ 					& $69.9\pm1.8$				 	& \CellR{$94.3\pm0.7$} 	& $73.5\pm1.7$ 					& -
							& $58.4\pm1.5$ 					& $48.8\pm4.0$					& \CellR{$90.7\pm1.2$} 	& $55.3\pm2.3$ 					& -											\\
    Cam3 			& $66.8\pm3.4$ 					& $72.5\pm2.6$					& $68.1\pm1.7$ 					& \CellR{$93.5\pm1.8$} 	& - 
							& $59.5\pm3.3$ 					& $51.4\pm3.0$					& $50.4\pm1.8$ 					& \CellR{$90.4\pm2.0$} 	& -											\\
    Cam4 			& - 										& -											& - 										& - 										& \CellR{$94.3\pm1.1$} 
							& - 										& -											& - 										& - 										& \CellR{$87.8\pm1.7$} 	\\
    \bottomrule
	\end{tabular}
	}
	\vspace{-1.5ex}
\end{table*}

In this work, 
we also evaluate the performance of the baseline methods under open-set scenario.
Specifically, 
the classifier is required to identify whether the given sample belongs to one of the known class given in the training set or rejects it as an unknown sample,
which is a realistic scenario in real-world applications.
Based on the findings on the above section, 
we only show the performance of IDT\_FV approach.
From \tab~\ref{tab:eval_result_SC} (in \textcolor{blue!80}{BLUE} background color) and \fig~\ref{fig:IDT_SC_3P},
the performance of IDT\_FV degraded on all scenarios.
In our preliminary experiment, 
we also evaluated the performance of open-set action recognition with KTH~\cite{Schuldt_ICPR_2004},
M$^2$I~\cite{Liu_TC_2016}, and
IXMAS~\cite{Weinland_ICCV_2007},
where similar performance trend are observed. 
Under MCAD, 
the most significant performance drop is with the {\it UseCellPhone} action, 
where the performance under all camera cases dropped from from 86.7\% to 45.1\%.

\begin{figure}[!t]
	\centering
	\begin{minipage}{1.0\columnwidth}
		\begin{minipage}{0.02\columnwidth}	\begin{sideways} \centerline{\footnotesize ~~~~~~Closed-Set Classification}\end{sideways} 	\end{minipage}
		\hfill
		\begin{minipage}{0.95\columnwidth}	\centerline{\includegraphics[width=1.0\linewidth]{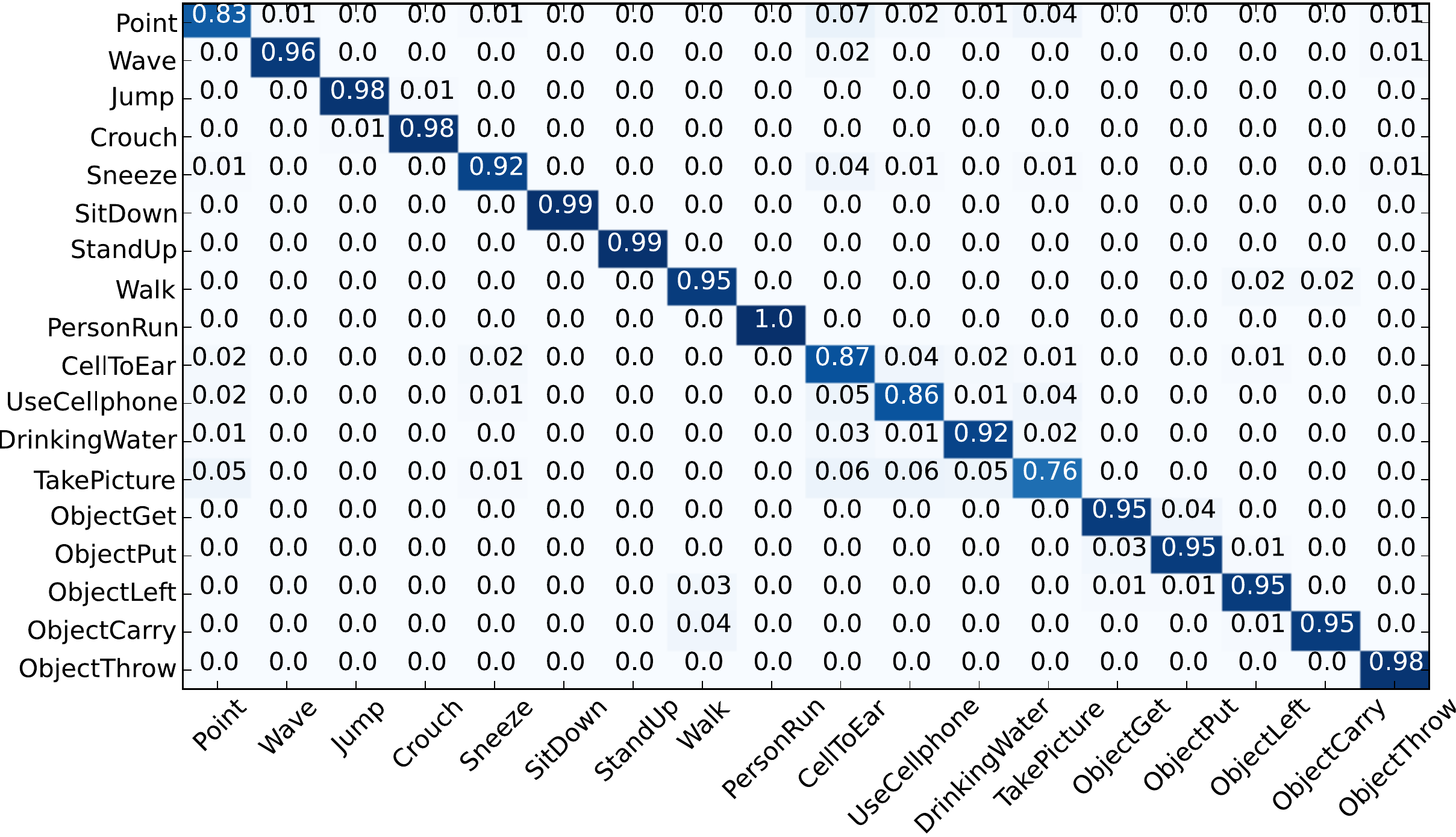}}	\end{minipage}
	\end{minipage}
	\begin{minipage}{1.0\columnwidth}
		\begin{minipage}{0.02\columnwidth}	\begin{sideways} \centerline{\footnotesize ~~~~~~Open-Set Classification}	\end{sideways} 	\end{minipage}
		\hfill
		\begin{minipage}{0.95\columnwidth}	\centerline{\includegraphics[width=1.0\linewidth]{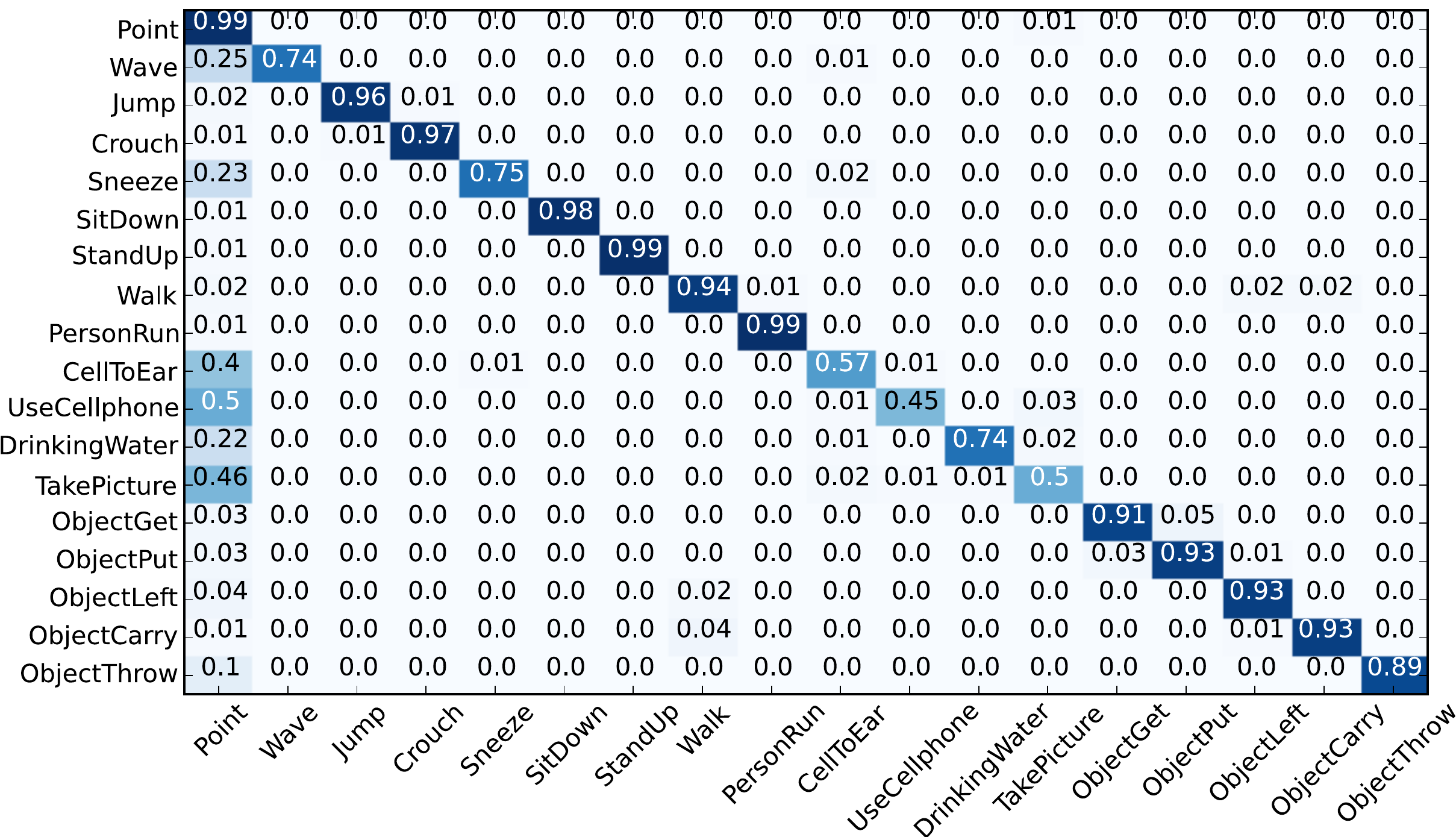}}	\end{minipage}
	\end{minipage}
	\vspace{-2ex}
	\caption
		{   
		\small
		Confusion matrix of IDT\_FV descriptor on the Evaluation set under AC scenario.  
		}
	\label{fig:SC_eval_confusion}
	\vspace{-2ex}
\end{figure}

Surprisingly, 
the performance of {\it Point} action increased for SC, ST, and AC scenarios.
To investigate this, 
we examine the confusion matrix of both closed-set and open-set scenario under AC scenario.
As shown in \fig~\ref{fig:SC_eval_confusion},
we find that the {\it Point} action is easier to confuse with other actions in the closed-set classification scenario,
where the probability of {\it Point} action to be misclassified as {\it CellToEar} action and {\it TakePicture} action are 0.07 and 0.04,
respectively.
While in the open-set scenario, 
the probability of {\it TakePicture} action reduced to 0.01.
We also observed that the performance of the micro actions with object 
(\ie~{\it CellToEar}, {\it UseCellphone}, {\it DrinkingWater}, and {\it TakePicture})
suffers significantly.
If we closely examine the confusion matrix, 
most of the test samples under these actions are mostly misclassified as {\it Point} action.
Similar to the {\it Point} action,
these action classes contain the arm motion action,
which might make it harder to distinguish with interest point based descriptor under the open-set scenario.  

\subsection{Evaluation with Open View Classification}
\label{sec:benchmark_open_view}

In this section,
we evaluate the benchmark performance for the open view recognition scenario,
\ie~the camera view of the test data has never been seen in the training phase.
We comprehensively evaluate all single camera cross view classification cases,
where data from one camera is selected to initialize the codebook and train the classifier (\ie~source view), 
and the evaluation is conducted on the selected camera (\ie~target view).
For all cases, 
the subjects in each data split is identical to those in Section~\ref{sec:benchmark_closed_view}.
We applied the optimal codebook size from the previous section and the SVM parameters are fine-tuned on the Development Set.
Both the close-set and open-set classification scenarios are evaluated.
The results with IDT\_FV on MCAD are shown in  \tab~\ref{tab:eval_result_SC_CC_2_IDT}.
Furthermore, 
we also conducted the same experiment on the synchronous IXMAS dataset (see \tab~\ref{tab:eval_result_SC_CC_2_IDT_IXMAS}).

Overall, 
the performance of cross view recognition drops significantly when compared to the single camera case.
The results are expected as the training and test data has significant difference in view perspective, FOV, image quality, and pixel resolution.
Consistent with the previous section, 
the performance with open-set classification method further reduce the performance.  
We also observed that the performance is more stable when the evaluated camera view has similar properties as that of the training data.
For instance, 
the Cam04-Cam05 pair report an average of 74.15\% on open-set classification scenario,
where the corresponding performance on Cam04-PTZ04 pair is around 75.4\%.
The view conditions of Cam06 is significantly different from the other cameras, 
and registers worst performance on all cross-view evaluation.  
In \tab~\ref{tab:eval_result_SC_CC_2_IDT_IXMAS},
we deliberately do not perform cross view evaluation for Cam4 because Cam4 is a top view camera (see \fig~\ref{fig:review_constraints}) and does not exhibit visually favorable properties for meaningful action recognition task. 

Finally,   
we highlight that the open view evaluation is essential to assess the robustness of any proposed algorithm.
Different from action recognition with consumer generated data
(\eg~egocentric video or crowdsourced dataset),
it is impractical to collect surveillance video data from all possible conditions for training purposes.
It is important to point out that for dataset that are synchronously recorded
(\eg~IXMAS dataset),
the evaluation needs to carefully designed such that temporal self-similarity is not utilized to improve the performance.
In our future work, 
we plan to evaluate view-invariant action recognition algorithms on the open view scenario.

\section{Conclusion}
\label{sec:conclusion}

In this paper, 
we presented a new action recognition dataset,
namely Multi-Camera Action Dataset (MCAD), 
which is designed to evaluate the open view action classification problem.
Different from existing multi-view datasets, 
the samples in MCAD are independently recorded with 5 cameras and 20 subjects, 
and contains a total of 14,298 action samples.
Inspired by the LFW dataset, 
we designed a standard evaluation protocol and benchmarked MCAD under several scenarios.

\section*{Acknowledgment}
\label{sec:acknowledgement}

This research is supported by the National Research Foundation, 
Prime Minister’s Office, 
Singapore under its International Research Centre in Singapore Funding Initiative.

{
\small
\bibliographystyle{ieee}
\bibliography{references}

\begin{thebibliography}{10}

\bibitem{ucf_aerial_2011}
{UCF} aerial action dataset.
\newblock http://server.cs.ucf.edu/ vision/aerial/index.html.

\bibitem{ucf_arg_2011}
{UCF} aerial camera, rooftop camera and ground camera dataset.
\newblock http://crcv.ucf.edu/data/UCF-ARG.php.

\bibitem{Aharon_TSP_2006}
M.~Aharon, M.~Elad, and A.~Bruckstein.
\newblock {K-SVD}: An algorithm for designing overcomplete dictionaries for
  sparse representation.
\newblock {\em {IEEE} Transactions on signal processing}, 54(11):4311--4322,
  2006.

\bibitem{Arthur_SDG_2007}
D.~Arthur and S.~Vassilvitskii.
\newblock k-means++: The advantages of careful seeding.
\newblock In {\em ACM-SIAM symposium on Discrete algorithms}, pages 1027--1035,
  2007.

\bibitem{Bach_ICML_2004}
F.~R. Bach, G.~R.~G. Lanckriet, and M.~I. Jordan.
\newblock Multiple kernel learning, conic duality, and the {SMO} algorithm.
\newblock In {\em ICML}, 2004.

\bibitem{Bendale_CVPR_2015}
A.~Bendale and T.~Boult.
\newblock Towards open world recognition.
\newblock In {\em CVPR}, pages 1893--1902, 2015.

\bibitem{Cui_TC_2014}
Z.~Cui, W.~Li, D.~Xu, S.~Shan, X.~Chen, and X.~Li.
\newblock Flowing on {R}iemannian manifold: Domain adaptation by shifting
  covariance.
\newblock {\em {IEEE} Transactions on Cybernetics}, 44(12):2264--2273, 2014.

\bibitem{Dollar_VSPETS_2005}
P.~Doll{\'a}r, V.~Rabaud, G.~Cottrell, and S.~Belongie.
\newblock Behavior recognition via sparse spatio-temporal features.
\newblock In {\em {IEEE} International Workshop on VS-PETS}, pages 65--72,
  2005.

\bibitem{Duan_PAMI_2012}
L.~Duan, I.~W. Tsang, and D.~Xu.
\newblock Domain transfer multiple kernel learning.
\newblock {\em {IEEE} Transactions on Pattern Analysis and Machine
  Intelligence}, 34(3):465--479, 2012.

\bibitem{Duan_PAMI_2012a}
L.~Duan, D.~Xu, I.~W. Tsang, and J.~Luo.
\newblock Visual event recognition in videos by learning from web data.
\newblock {\em {IEEE} Transactions on Pattern Analysis and Machine
  Intelligence}, 34(9):1667--1680, 2012.

\bibitem{Fan_JMLR_2008}
R.-E. Fan, K.-W. Chang, C.-J. Hsieh, X.-R. Wang, and C.-J. Lin.
\newblock {LIBLINEAR}: A library for large linear classification.
\newblock {\em Journal of Machine Learning Research}, 9:1871--1874, 2008.

\bibitem{Faraki_CV_2014}
M.~Faraki, M.~Palhang, and C.~Sanderson.
\newblock Log-{E}uclidean bag of words for human action recognition.
\newblock {\em {IET} Computer Vision}, 9(3):331--339, 2014.

\bibitem{Farhadi_LNCS_2008}
A.~Farhadi and M.~K. Tabrizi.
\newblock Learning to recognize activities from the wrong view point.
\newblock {\em Lecture Notes in Computer Science}, 5302:154--166, 2008.

\bibitem{Gao_NC_2016}
Z.~Gao, W.~Nie, A.~Liu, and H.~Zhang.
\newblock Evaluation of local spatial-temporal features for cross-view action
  recognition.
\newblock {\em Neurocomputing}, 173:110--117, 2016.

\bibitem{Gorelick_PAMI_2007}
L.~Gorelick, M.~Blank, E.~Shechtman, M.~Irani, and R.~Basri.
\newblock Actions as space-time shapes.
\newblock {\em {IEEE} Transactions on Pattern Analysis and Machine
  Intelligence}, 29(12):2247--2253, 2007.

\bibitem{Han_AFGR_2008}
L.~Han, W.~Liang, X.~Wu, and Y.~Jia.
\newblock Human action recognition using discriminative models in the learned
  hierarchical manifold space.
\newblock In {\em AFGR}, pages 1--6, 2008.

\bibitem{Chris_AVC_1998}
C.~Harris and M.~Stephens.
\newblock A combined corner and edge detector.
\newblock In {\em Alvey Vision Conference}, pages 23.1--23.6, 1988.

\bibitem{Hastie_springer_2001}
T.~Hastie, R.~Tibshirani, and J.~Friedman.
\newblock {\em The Elements of Statistical Learning: Data Mining, Inference,
  and Prediction}, volume~1.
\newblock Springer, 2001.

\bibitem{LFW_Tech_2007}
G.~B. Huang, M.~Ramesh, T.~Berg, and E.~Learned-Miller.
\newblock Labeled faces in the wild: A database for studying face recognition
  in unconstrained environments.
\newblock Technical report, Technical Report 07-49, University of
  Massachusetts, Amherst, 2007.

\bibitem{Hal_ACL_2007}
H.~D. III.
\newblock Frustratingly easy domain adaptation.
\newblock In {\em ACL}, 2007.

\bibitem{Jain_CVPR_2013}
A.~Jain, A.~Gupta, M.~Rodriguez, and L.~S. Davis.
\newblock Representing videos using mid-level discriminative patches.
\newblock In {\em CVPR}, pages 2571--2578, 2013.

\bibitem{Jiang_TC_2015}
Y.~Jiang, F.~Chung, S.~Wang, Z.~Deng, J.~Wang, and P.~Qian.
\newblock Collaborative fuzzy clustering from multiple weighted views.
\newblock {\em {IEEE} Transactions on Cybernetics}, 45(4):688--701, 2015.

\bibitem{yan_iccv_2007}
Y.~Ke, R.~Sukthankar, and M.~Hebert.
\newblock Event detection in crowded videos.
\newblock In {\em ICCV}, pages 1--8, 2007.

\bibitem{Kliper_PAMI_2012}
O.~Kliper-Gross, T.~Hassner, and L.~Wolf.
\newblock The action similarity labeling challenge.
\newblock {\em {IEEE} Transactions on Pattern Analysis and Machine
  Intelligence}, 34(3):615--621, 2012.

\bibitem{Kuehne_CVPR_2014}
H.~Kuehne, A.~Arslan, and T.~Serre.
\newblock The language of actions: Recovering the syntax and semantics of
  goal-directed human activities.
\newblock In {\em CVPR}, pages 780--787, 2014.

\bibitem{Kuehne_ICCV_2014}
H.~Kuehne, H.~Jhuang, E.~Garrote, T.~Poggio, and T.~Serre.
\newblock {HMDB}: a large video database for human motion recognition.
\newblock In {\em ICCV}, pages 2556--2563, 2011.

\bibitem{Laptev_ICCV_2003}
I.~Laptev and T.~Lindeberg.
\newblock Space-time interest points.
\newblock In {\em ICCV}, pages 432--439, 2003.

\bibitem{Laptev_CVPR_2008}
I.~Laptev, M.~Marszalek, C.~Schmid, and B.~Rozenfeld.
\newblock Learning realistic human actions from movies.
\newblock In {\em CVPR}, pages 1--8, 2008.

\bibitem{Liu_TC_2015}
A.-A. Liu, Y.-T. Su, P.-P. Jia, Z.~Gao, T.~Hao, and Z.-X. Yang.
\newblock Multipe/single-view human action recognition via part-induced
  multitask structural learning.
\newblock {\em {IEEE} Transactions on Cybernetics}, 45(6):1194--1208, 2015.

\bibitem{Liu_TC_2016}
A.-A. Liu, N.~Xu, W.-Z. Nie, Y.-T. Su, Y.~Wong, and M.~Kankanhalli.
\newblock Benchmarking a multi-modal \& multi-view \& interactive dataset for
  human action recognition.
\newblock {\em {IEEE} Transactions on Cybernetics}, 2016.
\newblock in press.

\bibitem{Liu_CVPR_2011}
J.~Liu, B.~Kuipers, and S.~Savarese.
\newblock Recognizing human actions by attributes.
\newblock In {\em CVPR}, pages 3337--3344, 2011.

\bibitem{Liu_CVPR_2009}
J.~Liu, J.~Luo, and M.~Shah.
\newblock Recognizing realistic actions from videos ``in the wild''.
\newblock In {\em CVPR}, pages 1996--2003, 2009.

\bibitem{Marszalek_CVPR_2009}
M.~Marszalek, I.~Laptev, and C.~Schmid.
\newblock Actions in context.
\newblock In {\em CVPR}, pages 2929--2936, 2009.

\bibitem{Messing_ICCV_2009}
R.~Messing, C.~Pal, and H.~Kautz.
\newblock Activity recognition using the velocity histories of tracked
  keypoints.
\newblock In {\em ICCV}, pages 104--111, 2009.

\bibitem{Niebles_LNCS_2010}
J.~C. Niebles, C.~Chen, and F.~Li.
\newblock Modeling temporal structure of decomposable motion segments for
  activity classification.
\newblock {\em Lecture Notes in Computer Science}, 6312:392--405, 2010.

\bibitem{Over_TRECVID_2014}
P.~Over, J.~Fiscus, G.~Sanders, D.~Joy, M.~Michel, G.~Awad, A.~Smeaton,
  W.~Kraaij, and G.~Qu{\'e}not.
\newblock {TRECVID} 2014 -- an overview of the goals, tasks, data, evaluation
  mechanisms and metrics.
\newblock In {\em Proceedings of TRECVID}, page~52, 2014.

\bibitem{Douglas_springer_2009}
D.~A. Reynolds.
\newblock Gaussian mixture models.
\newblock In {\em Encyclopedia of Biometrics}, pages 659--663. 2009.

\bibitem{Rodriguez_CVPR_2008}
M.~D. Rodriguez, J.~Ahmed, and M.~Shah.
\newblock Action {MACH} a spatio-temporal maximum average correlation height
  filter for action recognition.
\newblock In {\em CVPR}, 2008.

\bibitem{ryoo_iccv_2009}
M.~S. Ryoo and J.~K. Aggarwal.
\newblock Spatio-temporal relationship match: Video structure comparison for
  recognition of complex human activities.
\newblock In {\em ICCV}, pages 1593--1600, 2009.

\bibitem{Sapienza_IJCV_2014}
M.~Sapienza, F.~Cuzzolin, and P.~H. Torr.
\newblock Learning discriminative space-time action parts from weakly labelled
  videos.
\newblock {\em International Journal of Computer Vision}, 110(1):30--47, 2014.

\bibitem{Schuldt_ICPR_2004}
C.~Schuldt, I.~Laptev, and B.~Caputo.
\newblock Recognizing human actions: a local {SVM} approach.
\newblock In {\em ICPR}, pages 32--36, 2004.

\bibitem{Singh_AVSS_2010}
S.~Singh, S.~A. Velastin, and H.~Ragheb.
\newblock {MuHAVI}: A multicamera human action video dataset for the evaluation
  of action recognition methods.
\newblock In {\em AVSS}, pages 48--55, 2010.

\bibitem{Soomro_TR_2012}
K.~Soomro, A.~R. Zamir, and M.~Shah.
\newblock {UCF101}: {A} dataset of 101 human actions classes from videos in the
  wild.
\newblock Technical report, November 2012.

\bibitem{Tropp_TIT_2007}
J.~A. Tropp and A.~C. Gilbert.
\newblock Signal recovery from random measurements via orthogonal matching
  pursuit.
\newblock {\em {IEEE} Transactions on Information Theory}, 53(12):4655--4666,
  2007.

\bibitem{Wang_TSMCB_2012}
G.~Wang, F.~Wang, T.~Chen, D.~Yeung, and F.~H. Lochovsky.
\newblock Solution path for manifold regularized semisupervised classification.
\newblock {\em {IEEE} Transactions on Systems, Man, and Cybernetics, Part {B}},
  42(2):308--319, 2012.

\bibitem{Wang_IJCV_2013}
H.~Wang, A.~Kl{\"a}ser, C.~Schmid, and C.-L. Liu.
\newblock Dense trajectories and motion boundary descriptors for action
  recognition.
\newblock {\em International Journal of Computer Vision}, 103(1):60--79, 2013.

\bibitem{Wang_ICCV_2013}
H.~Wang and C.~Schmid.
\newblock Action recognition with improved trajectories.
\newblock In {\em ICCV}, pages 3551--3558, 2013.

\bibitem{Wang_ICCV_2013a}
L.~Wang, Y.~Qiao, and X.~Tang.
\newblock Mining motion atoms and phrases for complex action recognition.
\newblock In {\em ICCV}, pages 2680--2687, 2013.

\bibitem{Wang_TIP_2007}
L.~Wang and D.~Suter.
\newblock Learning and matching of dynamic shape manifolds for human action
  recognition.
\newblock {\em {IEEE} Transactions on Image Processing}, 16(6):1646--1661,
  2007.

\bibitem{Weinland_ICCV_2007}
D.~Weinland, E.~Boyer, and R.~Ronfard.
\newblock Action recognition from arbitrary views using 3{D} exemplars.
\newblock In {\em ICCV}, pages 1--7, 2007.

\bibitem{Xu_CVPR_2015}
Z.~Xu, Y.~Yang, and A.~G. Hauptmann.
\newblock A discriminative {CNN} video representation for event detection.
\newblock In {\em CVPR}, pages 1798--1807, 2015.

\bibitem{Yao_ICCV_2011}
B.~Yao, X.~Jiang, A.~Khosla, A.~L. Lin, L.~J. Guibas, and F.~Li.
\newblock Human action recognition by learning bases of action attributes and
  parts.
\newblock In {\em ICCV}, pages 1331--1338, 2011.

\bibitem{Zha_BMVC_2015}
S.~Zha, F.~Luisier, W.~Andrews, N.~Srivastava, and R.~Salakhutdinov.
\newblock Exploiting image-trained {CNN} architectures for unconstrained video
  classification.
\newblock In {\em BMVC}, pages 60.1--60.13, 2015.

\bibitem{Zhang_JEET_2015}
J.~Zhang, H.~Lin, W.~Nie, L.~Chaisorn, Y.~Wong, and M.~S. Kankanhalli.
\newblock Human action recognition bases on local action attributes.
\newblock {\em JEET}, 10(3), 2015.

\bibitem{Zheng_BMVC_2012}
J.~Zheng, Z.~Jiang, P.~J. Phillips, and R.~Chellappa.
\newblock Cross-view action recognition via a transferable dictionary pair.
\newblock In {\em BMVC}, pages 1--11, 2012.

\end{thebibliography}
}

\end{document}